\documentclass[sigconf, nonacm, screen]{acmart}%

\AtBeginDocument{%
  }
\usepackage{multirow}
\usepackage{threeparttable}
\usepackage{subcaption}
\usepackage{enumitem}

\setcopyright{acmlicensed}
\copyrightyear{2018}
\acmYear{2018}
\acmDOI{XXXXXXX.XXXXXXX}
\acmConference[Conference acronym 'XX]{Make sure to enter the correct
  conference title from your rights confirmation email}{June 03--05,
  2018}{Woodstock, NY}
\acmISBN{978-1-4503-XXXX-X/2018/06}

\begin{document}

\title{Cognitive Chunking for Soft Prompts: Accelerating Compressor Learning via Block-wise Causal Masking}

\author{Guojie Liu}
\authornote{Both authors contributed equally to this research.}
\email{liuguojie@nudt.edu.cn}
\author{Yiqi Wang}
\authornotemark[1]
\email{yiq@nudt.edu.cn}
\affiliation{%
  \institution{National University of Defense Technology}
  \city{Changsha}
  \country{China}
}

\author{Yanfeng Yang}
\affiliation{%
  \institution{National University of Defense Technology}
  \city{Changsha}
  \country{China}}
\email{yanfengyang@nudt.edu.cn}

\author{Wenqi Fan}
\affiliation{%
  \institution{The Hong Kong Polytechnic University}
  \city{Hong Kong SAR}
  \country{China}
}
\email{wenqifan03@gmail.com}

\author{Songlei Jian}
\affiliation{%
 \institution{National University of Defense Technology}
 \city{Changsha}
 \country{China}}
\email{jiansonglei@nudt.edu.cn}

\author{Jianfeng Zhang}
\affiliation{%
  \institution{National University of Defense Technology}
  \city{Changsha}
  \country{China}}
\email{jfzhang@nudt.edu.cn}

\author{Jie Yu}
\affiliation{%
  \institution{National University of Defense Technology}
  \city{Changsha}
  \country{China}}
\email{yj@nudt.edu.cn}

\renewcommand{\shortauthors}{Trovato et al.}
\newcommand{\yq}[1]{%
  \textcolor{red}{\textbf{[YQ: #1]}}%
}
\newcommand{\wq}[1]{%
  \textcolor{red}{\textbf{[wq: #1]}}%
}
\newcommand{\gj}[1]{%
  \textcolor{blue}{\textbf{[GJ: #1]}}%
}
\newcommand{\sys}{\textsc{PIC}}
\newcommand\SystemName{\sys}

\begin{abstract}
Providing extensive context via prompting is vital for leveraging the capabilities of Large Language Models (LLMs). However, lengthy contexts significantly increase inference latency, as the computational cost of self-attention grows quadratically with sequence length. To mitigate this issue, context compression—particularly soft prompt compression—has emerged as a widely studied solution, which converts long contexts into shorter memory embeddings via a trained compressor. Existing methods typically compress the entire context indiscriminately into a set of memory tokens, requiring the compressor to capture global dependencies and necessitating extensive pre-training data to learn effective patterns.
Inspired by the chunking mechanism in human working memory and empirical observations of the spatial specialization of memory embeddings relative to original tokens, we propose Parallelized Iterative Compression (\sys). By simply modifying the Transformer's attention mask, \sys~explicitly restricts the receptive field of memory tokens to sequential local chunks, thereby lowering the difficulty of compressor training.
Experiments across multiple downstream tasks demonstrate that \sys~consistently outperforms competitive baselines, with superiority being particularly pronounced in high compression scenarios (e.g., achieving relative improvements of 29.8\% in F1 score and 40.7\% in EM score on QA tasks at the $64\times$ compression ratio). Furthermore, \sys~significantly expedites the training process. Specifically, when training the 16$\times$ compressor, it surpasses the peak performance of the competitive baseline while effectively reducing the training time by approximately 40\%.

\end{abstract}

\keywords{Context Compression, Large Language Models, Soft Prompt, Memory Embeddings}

\maketitle

\section{Introduction}
To enhance the generation quality of Large Language Models (LLMs), incorporating extensive contextual information has become a common practice, providing essential background and task-specific clues in paradigms such as Retrieval-Augmented Generation (RAG) \cite{gao2023retrieval} and In-Context Learning (ICL) \cite{li2024long}. However, the extension of context length inevitably introduces significant computational overhead, leading to a substantial increase in inference latency and posing critical challenges for real-time applications.

A promising direction to mitigate these issues is context compression, which aims to reduce the effective length of the input sequence while preserving its informational core. Existing compression approaches can be broadly categorized into two types: hard prompt compression and soft prompt compression~\cite{li2025prompt}. Hard prompt compression typically relies on heuristic rules to prune redundant tokens from the original context or employs another LLM to produce a discrete, natural language summary, fundamentally retaining a symbolic representation~\cite{li2023compressing, jiang2023llmlingua, chuang2024learning, pan2024llmlingua, jiang2024longllmlingua}.
In contrast, soft prompt compression learns a compact, continuous representation by distilling the context information into a small set of dense embedding vectors~\cite{chevalier2023adapting, ge2024incontext, 10.1145/3701551.3703527, cheng2024xrag, louis-etal-2025-pisco, dai2025pretraining}, which can be viewed as a novel modality offering the potential for higher compression rates \cite{dai2025pretraining}. Therefore, this work focuses on soft prompt compression. Typically, the input context is processed by a compressor to yield a shortened sequence of continuous memory embeddings. These compressed representations—often further aligned in dimension and semantics via a converter layer—then replace the original context as input to the downstream decoder for generation.
\begin{figure}
    \centering
    \includegraphics[width=1\linewidth]{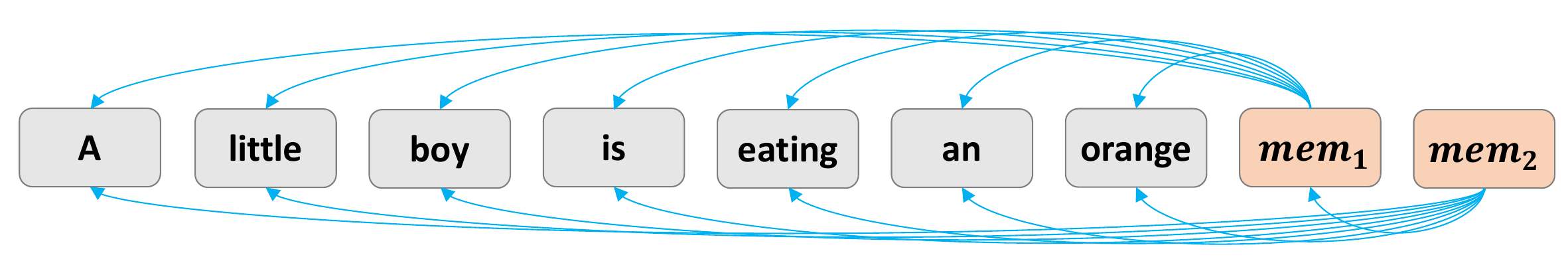}
    \caption{An illustration of unconstrained global attention in a typical soft prompt compression method: both memory tokens need to attend to all the input  words simultaneously.}
    \Description{Each memory token (e.g., $mem_1, mem_2$) is required to attend to the entire input sequence simultaneously. This lack of structural constraint imposes a heavy burden on the compressor to implicitly learn information routing and allocation from scratch.}
    \label{fig.1}
\end{figure}

However, the majority of existing methods typically follow the described pipeline yet rely on unconstrained global attention to map the entire context into the fixed memory slots, as shown in Figure \ref{fig.1}. 
In this design, the compressor lacks an inductive bias to establish explicit correspondences between specific memory tokens and particular input subsequences. As a result, each memory token typically needs to attend to and extract information from all tokens in the lengthy context~\cite{yang2018modeling}, which encourages the model to implicitly learn complex global routing patterns. This process often increases the learning difficulty, tending to render the training process more data-intensive and time-consuming.~\cite{goyal2022inductive}

\begin{figure}
    \centering
    \captionsetup[subfigure]{skip=2pt} 
    \captionsetup[subfigure]{font=normalsize}
    \begin{subfigure}{0.9\linewidth}
        \centering
        \includegraphics[width=\textwidth]{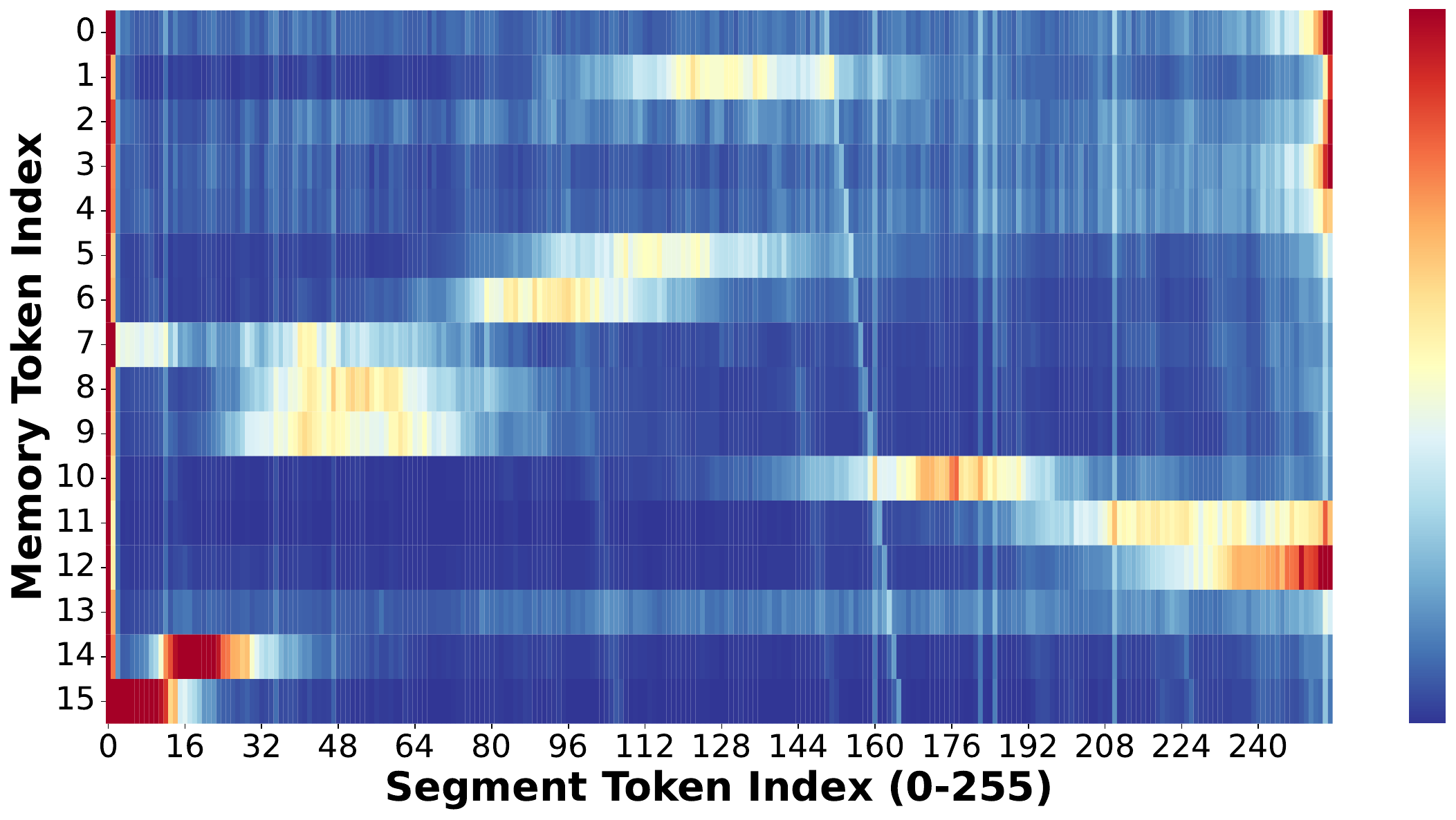}
        \caption{\textbf{Compressor Trained for 1k Steps}}
        \label{fig.2a}
    \end{subfigure}
    
    \vspace{2pt} 
    
    \begin{subfigure}{0.9\linewidth}
        \centering
        \includegraphics[width=\textwidth]{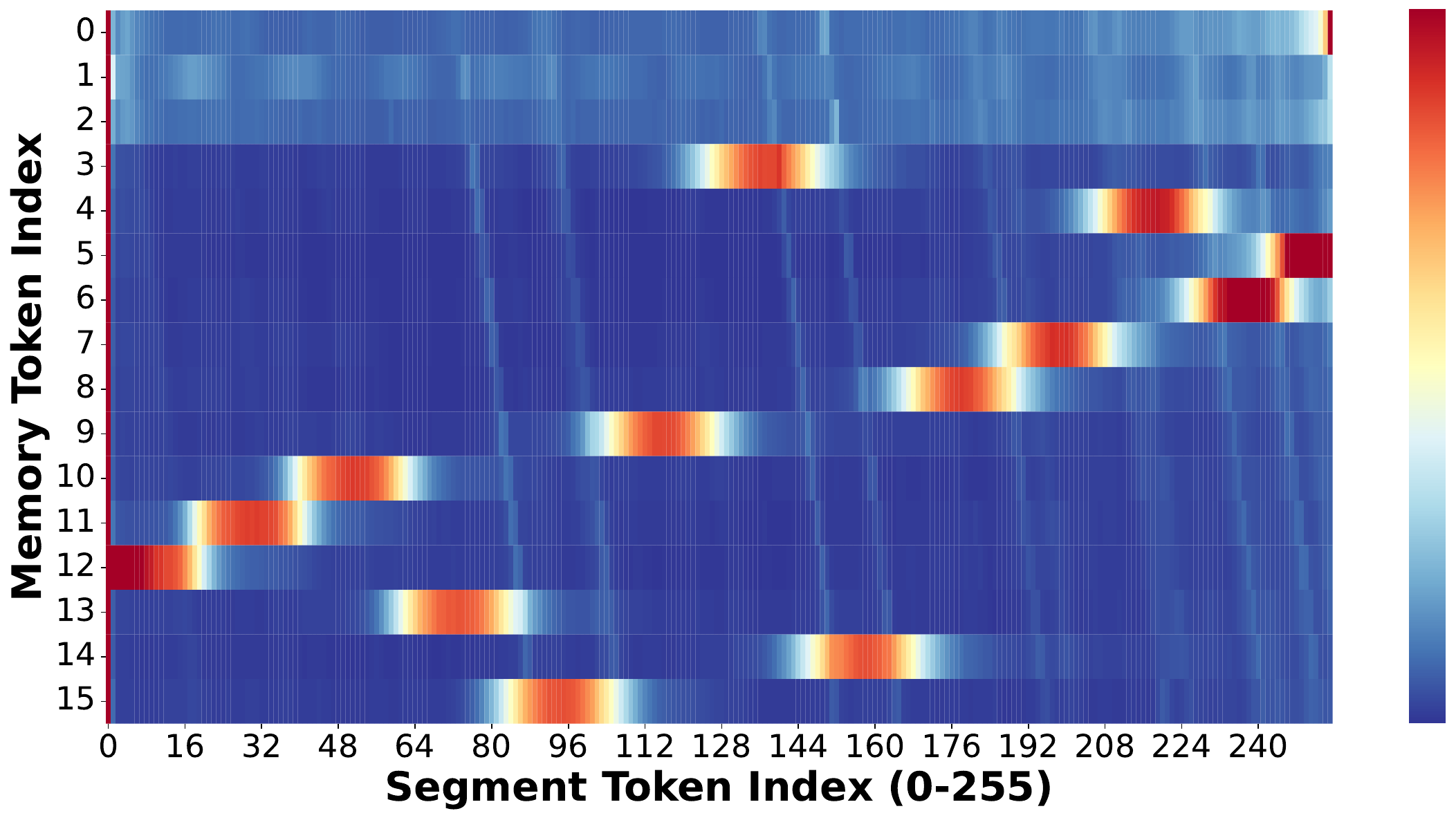}
        \caption{\textbf{Fully Trained Compressor (90k Steps)}}
        \label{fig.2b}
    \end{subfigure}
    
    \setlength{\abovecaptionskip}{5pt} 
    \caption{Attention weight heatmap between memory tokens and original context tokens. Red indicates higher attention weight, while blue represents lower attention weight.}

    \label{fig.2}
\end{figure}

\begin{figure}
    \centering
    \includegraphics[width=0.9\linewidth]{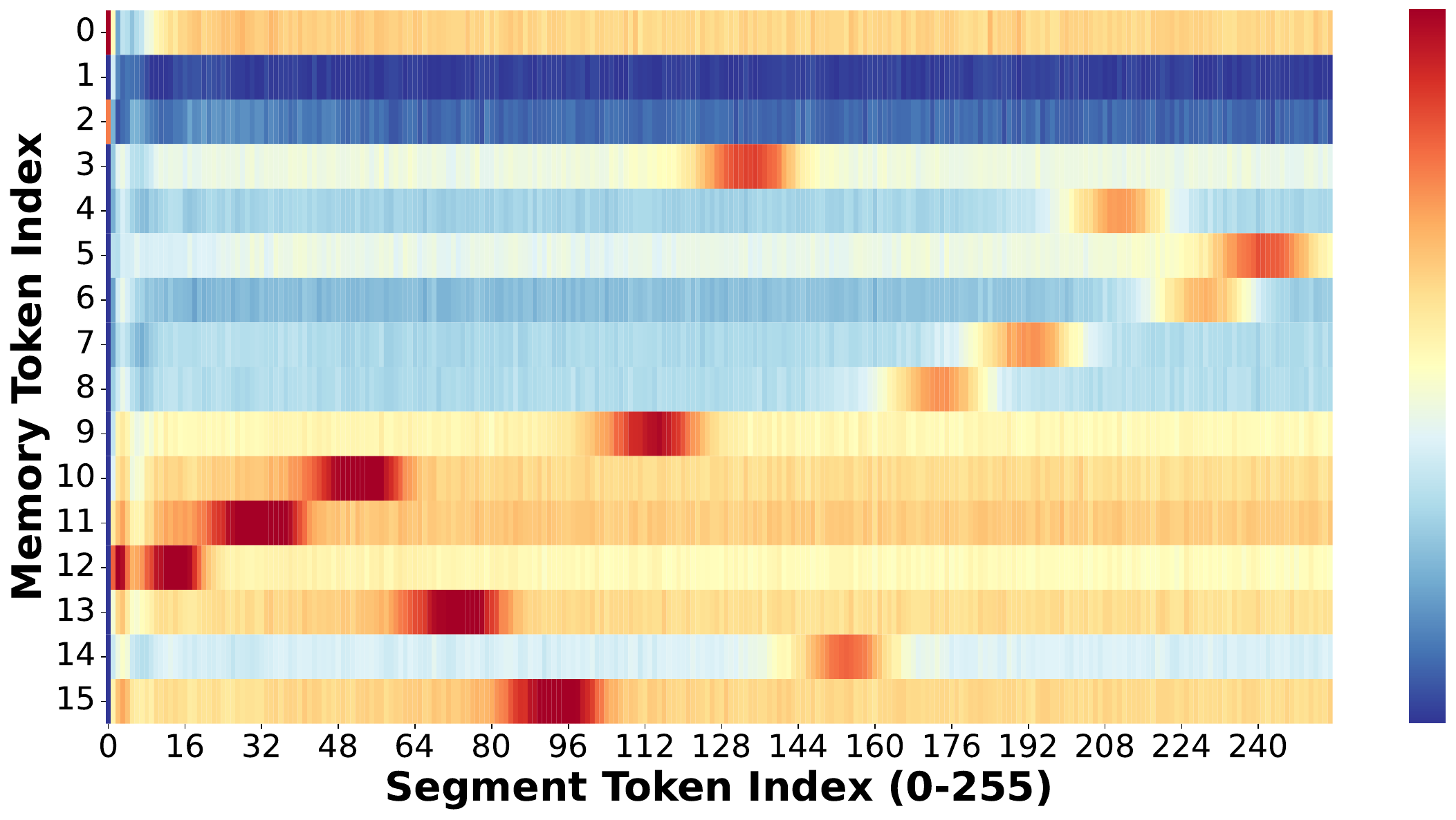}
        \vskip -0.15in
    \caption{Heatmap visualization of the cosine similarity between generated memory embeddings generated by the fully trained compressor (90k steps) and the embeddings of original context tokens on the HotpotQA dataset. Red indicates higher similarity, while blue represents lower similarity.}

    \label{fig.3}
\end{figure}

The core challenge, therefore, lies in how the compressor allocates its limited memory tokens to retain essential information. To demystify the learned allocation strategy, we examine the interaction between memory tokens and original context tokens using a representative framework~\cite{dai2025pretraining}. 
We compute both the attention weights and the cosine similarity between their corresponding embeddings on the HotpotQA dataset. The resulting heatmaps in Figure \ref{fig.2} (attention) and Figure \ref{fig.3} (embedding similarity) reveal a highly consistent and crucial evolution. As shown in Figure \ref{fig.2a}, early in training (1k steps), attention from memory tokens to the input is diffuse and unstructured. However, after full convergence (90k steps), a clear block-wise pattern emerges (Figure \ref{fig.2b}), where each memory token aligns strongly with a specific, contiguous segment of the context. This spatial specialization is not an artifact of the attention mechanism alone; it is directly corroborated by the embedding similarities in Figure 3, which exhibit an identical block-wise structure. 
This spatial specialization suggests that an desired strategy for maximizing information retention under the memory bottleneck is to partition the context and dedicate individual memory tokens to distinct blocks.

This observed correspondence, where memory tokens become selectively attuned to contiguous segments of the input context, bears a notable resemblance to the cognitive principle of chunking in human working memory~\cite{baddeley1992working, barak2014working}. It is evident from cognitive studies that working memory, constrained by a limited capacity~\cite{cowan2010magical}, compensates by organizing information into coherent, higher-order units or "chunks"—such as grouping digits in a phone number or phrases in a sentence—thereby facilitating efficient storage and recall~\cite{zhong2024synaptic}. However, a critical distinction exists: human comprehension unfolds sequentially, with each new segment being interpreted in light of previously integrated information. In contrast, existing compressors acquire this partitioning strategy implicitly, without enforcing a strict sequential correspondence between memory tokens and text blocks. This observation motivates us to explicitly impose a sequential, block-wise alignment, guiding each memory token to attend to a specific, ordered segment of the context. By doing so, we simplify the learning objective from global dependency modeling to localized information extraction, which has the potential to substantially reducing optimization complexity. To preserve the efficiency of parallel computation while maintaining sequential integrity, we further introduce a Parallelized Iterative Compression (\sys) approach. It employs a block-wise causal mask that simulates incremental, block-by-block processing in a single forward pass, thereby preserving order constraints without incurring the latency of sequential iteration.

Experiments on two representative downstream tasks—namely Retrieval-Augmented Generation (RAG) based Question Answering and In‑Context Learning (ICL)—demonstrate that our method consistently outperforms competitive baselines, with the performance gap widening as the compression ratio increases. While achieving competitive results at low compression rates, our approach exhibits significantly stronger robustness when the memory budget becomes severely constrained (e.g., under 16$\times$ and 64$\times$ compression), confirming its superior efficiency in preserving task‑relevant information. To further verify that our design effectively reduces learning difficulty, we conduct a data‑efficiency analysis. The results show that our method is substantially more data-efficient; specifically, under the 16$\times$ compression setting, it enables the model to surpass the peak performance of baselines with approximately 40\% less training time.
The key contributions of this paper are summarized as follows:
\begin{itemize}[leftmargin=*]
    \item Through a systematic investigation of the interaction between memory embeddings and input context,
    we identify a clear spatial‑specialization phenomenon that closely mirrors the cognitive chunking mechanism in human working memory.
    
    \item 
    We propose an iterative compression method that explicitly restricts each memory token to attend only to a sequential block of the input context, thereby simplifying global modeling into local extraction.

    \item To enable efficient parallel execution while preserving sequential logic, we introduce Parallelized Iterative Compression (\sys), which approximates incremental block‑wise processing via a novel block‑wise causal mask.
    
    \item Extensive experiments demonstrate that our method not only accelerates compressor convergence but also achieves superior performance across multiple downstream tasks. Visualization studies further confirm that our approach successfully enforces the intended sequential block‑to‑token alignment and yields more orthogonal representations across different memory tokens.
\end{itemize}

\section{Related Work}
Existing context compression approaches fall into two broad categories: hard prompt compression and soft prompt compressionn~\cite{li2025prompt}. Hard prompt compression, such as LLMLingua2~\cite{pan2024llmlingua}, produces compressed outputs in human-readable natural language, often by formulating the task as token-level classification. Our work focuses on soft prompt compression, which condenses text into continuous memory representations. Early methods like AutoCompressor~\cite{chevalier2023adapting} introduced iterative summary generation, while PPC~\cite{mu2023learning} demonstrated that simple attention masks could guide models to compress prompts into “gist tokens.” Some follow-up works, such as xRAG~\cite{cheng2024xrag}, employ knowledge distillation as their primary training strategy, focusing on training only the modality bridge. Other frameworks, including ICAE~\cite{ge2024incontext}, present a general compressor-decoder architecture pre-trained via reconstruction and prediction tasks. COCOM~\cite{10.1145/3701551.3703527} further emphasizes that the downstream decoder requires additional fine-tuning to effectively utilize compressed memory. Subsequent analyses, such as PISCO~\cite{louis-etal-2025-pisco}, examine embedding behaviors in such settings. Building on these, PCC~\cite{dai2025pretraining} explores fundamental design questions like scalability across compression ratios.

\section{Method}
Standard global compression approaches, which process the context indiscriminately, impose a heavy burden on the compressor to implicitly learn complex attention patterns. Motivated by the spatial specialization observed and the sequential nature of human reading, we introduce a sequential block-wise paradigm as the foundational logic for effective compression. This paradigm explicitly mandates that context is compressed into memory slots through structured sequential segments to reduce optimization difficulty. Although a direct iterative implementation of this logic ensures structural integrity, it suffers from high latency. To resolve this trade-off between structure and efficiency, we propose Parallelized Iterative Compression (\sys). By leveraging a block-wise causal mask, \sys~approximates the sequential iterative process in a fully parallelized manner. This design effectively retains the inductive bias of sequential processing while achieving the inference speed characteristic of direct global compression.

\subsection{Problem Definition}
The objective of soft prompt compression is to distill an input context $\mathbf{X} = (x_1, \dots, x_L)$ of $L$ tokens into a compact memory representation $\widetilde{\mathbf{H}} = (h_1, \dots, h_N)$, which is a concatenation of $N$ embedding vectors, $N \ll L$ and $L$ is divisible by $N$.
These compressed embeddings subsequently replace the original context $\mathbf{X}$ to facilitate downstream text generation tasks in an LLM:
\begin{equation}
    f_{\text{LLM}}(\widetilde{\mathbf{H}}, \text{prompt}) \rightarrow \mathbf{Y}_{\text{output}}
\end{equation}

\begin{figure*}
    \centering
    \includegraphics[width=1\linewidth]{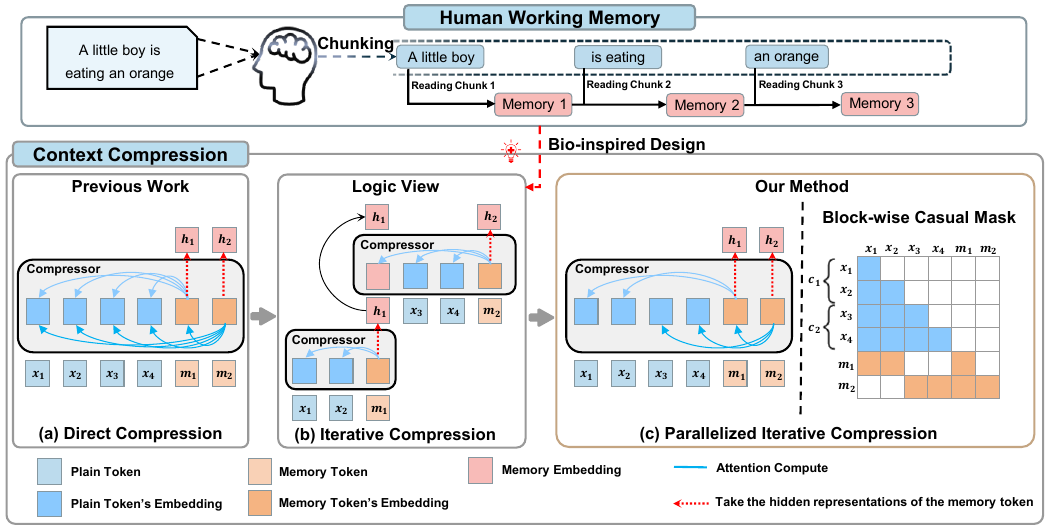}
    \caption{Comparison of different compression paradigms. (a) Direct Compression: Each memory token attends globally to the entire input context without structural constraints. (b) Iterative Compression: Inspired by the chunking mechanism in human working memory, we split the input context into chunks, and memory tokens are generated sequentially, with each token attending only to its specific chunk and previous memory. (c) Parallelized Iterative Compression (\sys): Our proposed method allows all memory tokens to be processed in parallel within a single sequence, using a block-wise causal mask to enforce the same local dependency constraints as the iterative approach. }
    \label{fig.4}
\end{figure*}

Most of the existent compression methods, as illustrated in Figure \ref{fig.4}(a), formulate compression as a global mapping problem~\cite{dai2025pretraining}. By introducing a set of $N$ learnable memory tokens $\widetilde{\mathbf{M}} = (m_1, \dots, m_N)$ as placeholders, the compressor processes the context $\mathbf{X}$ along with these memory tokens to obtain the hidden states from the final layer as the memory embeddings $\widetilde{\mathbf{H}}$. Therefore, the generation of each memory embedding $h_t$ depends on the entire input context $\mathbf{X}$ and the preceding memory tokens $m_{<t}$:
\begin{equation}
    h_t = \text{Compressor}(\mathbf{X}, m_{<t})
\end{equation}
However, this direct compression approach can necessitate learning complex attention patterns and is typically sub-optimal.

\subsection{Iterative Compression}
To mitigate the difficulty of learning global dependencies, we draw inspiration from the limit capacity of human working memory and adopt an iterative compression strategy. As visually depicted in Figure \ref{fig.4}, human working memory employs a chunking mechanism to organize continuous streams of information into discrete, manageable units for efficient processing. Mimicking this mechanism as shown in Figure \ref{fig.4}(b), we decompose the input $\mathbf{X}$ into $N$ equal-length non-overlapping chunks $\mathbf{X} = (c_1, c_2, \dots, c_N)$.  
The compression process is then structured sequentially, where the $t$-th memory embedding $h_t$ is generated by compressing only its corresponding chunk $c_t$, while maintaining access to the historical compressed memory $h_{<t}$. Accordingly, the generation process is reformulated as:
\begin{equation}
    h_t = \text{Compressor}(c_t, h_{<t})
\end{equation}
This formulation introduces a structural inductive bias, enforcing a one-to-one mapping between memory tokens and context chunks. Consequently, the generation of $h_t$ becomes local to the original text (attending only to $c_t$) but global to the compressed memory history ($h_{<t}$).

While iterative decomposition effectively simplifies global modeling by breaking it down into manageable local tasks, thereby alleviating the training burden, its strict sequential processing dependency inherently prohibits parallel computation. As a consequence, inference latency increases linearly with the number of memory tokens, creating a fundamental trade-off between optimization tractability and computational efficiency.
\subsection{Parallelized Iterative Compression}
To resolve the latency bottleneck of iterative compression while preserving its beneficial learning properties, we propose Parallelized Iterative Compression (\sys).
As illustrated in Figure \ref{fig.4}(c), \sys~processes all context chunks and memory tokens simultaneously in a single forward pass. Crucially, it employs a block-wise causal attention mask to strictly simulate the receptive field of the sequential iterative process. This design allows the model to learn the simplified local objectives of iterative compression without incurring the latency of serial execution.

\subsubsection{Sequence Construction}
Unlike serial iterative methods that process chunks step-by-step, \sys~concatenates all chunks and memory tokens into a single sequence for parallel processing. Let $\mathbf{X}$ be uniformly divided into $N$ chunks $\mathbf{X} = (c_1, c_2, \dots, c_N)$, and $\widetilde{\mathbf{M}} = (m_1, \dots, m_N)$ be the appended memory placeholders. The input sequence $\mathbf{Z}$ is constructed as:
\begin{equation}
    \mathbf{Z} = [c_1, c_2, \dots, c_N, m_1, m_2, \dots, m_N] = [\mathbf{X}, \widetilde{\mathbf{M}}]
\end{equation}
\subsubsection{Block-wise Causal Attention Mask}
Our core innovation is the block-wise causal attention mask, which enforces the iterative inductive bias within a single parallel pass. We design the mask $\mathbf{M} \in \mathbb{R}^{|\mathbf{Z}| \times |\mathbf{Z}|}$ to ensure that each memory token $m_t$ attends exclusively to its assigned chunk $c_t$ and the memory history $m_{<t}$.

Let $z_i, z_j$ be the query and key tokens in $\mathbf{Z}$. We define a binary visibility indicator function $\text{Visible}(i, j) \in \{0, 1\}$, where 1 denotes that $z_i$ is allowed to attend to $z_j$, according to three rules:
\begin{enumerate}
    \item \textbf{Intra-Context Causal:} For tokens within the context ($z_i, z_j \in \mathbf{X}$), $\text{Visible}(i, j) = 1$ if $i \ge j$.
    \item \textbf{Intra-Memory Causal:} For tokens within the memory ($z_i, z_j \in \widetilde{\mathbf{M}}$), $\text{Visible}(i, j) = 1$ if $i \ge j$.
    \item \textbf{Memory-to-Block Attention:} If the query is a memory token ($z_i \in \widetilde{\mathbf{M}}$ at index $t$) and the key is a context token ($z_j \in \mathbf{X}$ belonging to chunk $c_k$), $\text{Visible}(i, j) = 1$ if $t = k$.
\end{enumerate}

Formally, the mask entry $M_{i,j}$ is:
\begin{equation}\label{eq.5}
    M_{i,j} = \begin{cases} 
        0 & \text{if } \text{Visible}(i, j) \\
        -\infty & \text{otherwise}
    \end{cases}
\end{equation}

This design effectively prevents $m_t$ from accessing past chunks $c_{<t}$ directly, forcing the compressor to rely on the memory history $m_{<t}$ for context coherence. By compressing locally and connecting globally, \sys~achieves logical equivalence to iterative compression without the serial latency.

\subsection{Pre-training and Fine-tuning}
To ensure a fair comparison and rigorously validate the effectiveness of our block-wise causal mask, we align our training pipeline with the established PCC framework~\cite{dai2025pretraining}, utilizing identical pre-training tasks and fine-tuning objectives.

\subsubsection{Pre-training Objectives}
\noindent \textbf{Text Reconstruction (TR).} This task requires the model to reconstruct the original context $\mathbf{X}$ auto-regressively, conditioned on the compressed memory $\widetilde{\mathbf{H}}$. A special token \texttt{<AE>} is appended to $\widetilde{\mathbf{H}}$ to indicate the reconstruction task. The objective is formalized as:
\begin{equation}
    \mathcal{L}_{\text{TR}} = - \frac{1}{L} \sum_{i=1}^{L} \log f_{\text{LLM}}(x_i \mid \widetilde{\mathbf{H}}, \texttt{<AE>}, x_{<i})
\end{equation}
where $x_{<i}$ denotes the prefix sequence $(x_1, \dots, x_{i-1})$.

\noindent \textbf{Text Completion (TC).} This objective encourages the memory representation to capture semantic information essential for future generation. We partition the text into a prefix and a continuation, where the memory $\widetilde{\mathbf{H}}$ is compressed from the prefix. The model then predicts the continuation sequence starting from index $k+1$:
\begin{equation}
    \mathcal{L}_{\text{TC}} = - \frac{1}{L-k} \sum_{i=k+1}^{L} \log f_{\text{LLM}}(x_i \mid \widetilde{\mathbf{H}}, x_{k}, \dots, x_{i-1})
\end{equation}
The total pre-training loss is a weighted sum of both objectives, with $\lambda = 0.5$:
\begin{equation}
    \mathcal{L} = \lambda \mathcal{L}_{\text{TC}} + (1 - \lambda) \mathcal{L}_{\text{TR}}
\end{equation}

\subsubsection{Fine-tuning}
For downstream adaptation, we conduct domain-specific fine-tuning. Following the experimental settings in PCC, we utilize a single representative dataset per task category to adapt the compressor.

\section{Experiments}

\subsection{Experimental Setup}

\subsubsection{Datasets}

Table \ref{tab.1} provides a summary of the datasets utilized across different training stages. In the pre-training stage, we leveraged approximately 3 billion tokens from the FineWeb~\cite{penedo2024fineweb} dataset. For the fine-tuning stage, two datasets were incorporated: SQuAD~\cite{rajpurkar2018know}, comprising 86,821 training samples and 5,928 test samples, and GSM8K~\cite{cobbe2021training}, containing 6,725 training samples and 748 test samples. Specifically, We employed the training set of SQuAD as the domain-specific dataset for the RAG-based Question Answering (QA) task, and GSM8K for the In-Context Learning (ICL) task.
\begin{table}
    \centering
    \caption{Basic Statistics of Data Samples for Datasets}
    \label{tab.1}
    \begin{tabular*}{\linewidth}{@{\extracolsep{\fill}}cccc} 
        \toprule
        Stage & Source & Train & Test\\
        \midrule
        Pre-training & FineWeb & 11,520,000 & 576\\
        Fine-tuning(RAG) & SQuAD & 86,821 & 5,928\\
        Fine-tuning(ICL) & GSM8K & 6,725 & 748\\
        \bottomrule
    \end{tabular*}
\end{table}

\subsubsection{Implementation Details}
Our experiments primarily focus on efficient context compression using accessible model scales. During the pre-training phase, we explored architectures from two different model families as compressors. Primarily, we engaged Qwen2.5-0.5B-Instruct as the compressor, trained via full-parameter fine-tuning, while the downstream decoder was Llama-3-8B-Instruct, which remained frozen. To eliminate potential biases arising from model family differences, we additionally trained a compressor based on Llama-3.2-1B-Instruct, also using full-parameter fine-tuning.
In the pre-training stage, the learning rate was set to $1 \times 10^{-4}$ for all models. Training the \sys~ model (with the Qwen backbone) for one full epoch on the pre-training dataset took approximately 170 hours on a single node with two 48GB RTX 4090 GPUs. For all inference experiments, greedy decoding was applied to the LLM decoder.

\subsubsection{Baselines}
We compare our approach against six competitive context compression methods:

\noindent\textbf{AutoCompressor}~\cite{chevalier2023adapting}: Adapts a pre-trained Llama-2-7b-hf to generate summaries embeddings, achieving a high compression ratio of 40x.

\noindent\textbf{xRAG}~\cite{cheng2024xrag}: Utilizes Mistral-7B-Instruct-v0.2 as the downstream LLM. It focuses on training the connector (Convert module) while keeping other components frozen, fine-tuned on multiple downstream datasets.

\noindent\textbf{COCOM}~\cite{10.1145/3701551.3703527}: Employs a compressor based on bert-base-uncased for pre-training and fine-tunes the downstream LLM (Mistral-7B-Instruct-v0.2).

\noindent\textbf{ICAE}~\cite{ge2024incontext}: Features a compressor pre-trained on Mistral-7B to compress context into memory slots, utilizing Mistral-7B-Instruct-v0.2 as the downstream LLM.

\noindent\textbf{LLMLingua-2}~\cite{pan2024llmlingua}: A hard prompt compression method based on token classification (using XLM-RoBERTa-large). It is evaluated with Llama-3-8B-Instruct as the target LLM.

\noindent\textbf{PCC}~\cite{dai2025pretraining}: The primary baseline. PCC-Lite utilizes a compressor pre-trained on GPT2-Large, while PCC-Large employs Llama-3-8B-Instruct. Both variants use Llama-3-8B-Instruct as the downstream decoder.

To ensure a fair comparison, we strictly adhered to the experimental settings established in PCC and directly compared our results against the metrics reported in their paper.

\subsection{Main Results}

\subsubsection{RAG-based Question Answering}
We first evaluate our compressor in the Retrieval-Augmented Generation (RAG) based Question Answering scenario. Each data sample consists of a context, a question, and an corresponding answer. We perform domain-specific fine-tuning on the pre-trained compressor using the SQuAD training set, followed by evaluation on SQuAD, HotpotQA \cite{yang2018hotpotqa}, AdversarialQA \cite{bartolo2020beat}, and Natural Questions (NQ) \cite{karpukhin2020dense}. \textbf{Reference (w/o Context)} refers to generating responses using the downstream decoder conditioned solely on the question, without any additional context. \textbf{Reference (w/ Context)} denotes concatenating the uncompressed original context with the question as the input prompt.

For baseline methods, we maintain consistency with the settings in the PCC paper. To enable a direct and fair comparison, we utilize identical evaluation scripts and assess performance using F1 score and Exact Match (EM) metrics. The F1 score represents the harmonic mean of precision and recall, while the EM score indicates whether the generated response is identical to the ground truth answer.

The results are presented in Table \ref{tab.2}. On average, our method consistently surpasses both PCC-Lite and PCC-Large across all compression ratios. 
Notably, the performance advantage of our method becomes increasingly pronounced as the memory token budget decreases (e.g., at 16x and 64x compression).
This empirical evidence strongly validates the efficiency of our sequential blocking strategy in maximizing information retention under constrained memory conditions.

\begin{table*}[h]
    \centering
    \caption{Comparative performance of various compression methods across four QA datasets. Bold values indicate the best performance for each dataset within the same compression ratio category, and second-best results are underlined. The values in parentheses denote the parameter size of the compressor used by each method.}
    
    \vspace{2mm}
    \begin{threeparttable} 
    \resizebox{\textwidth}{!}{
    \begin{tabular}{ccccccccccccc}
        \toprule
        \multirow{2}{*}{Compression Rate} & \multirow{2}{*}{Methods} & \multicolumn{2}{c}{SQuAD} & \multicolumn{2}{c}{HotPotQA} & \multicolumn{2}{c}{AdversarialQA} & \multicolumn{2}{c}{NQ} & \multicolumn{2}{c}{Average}\\
        \cmidrule(lr){3-4} \cmidrule(lr){5-6} \cmidrule(lr){7-8} \cmidrule(lr){9-10} \cmidrule(lr){11-12}
          &  & $F_1$ & EM & $F_1$ & EM & $F_1$ & EM & $F_1$ & EM & $F_1$ & EM &\\
        \midrule
        \multirow{2}{*}{$1\times$} & w/o Context & 14.84   & 3.41    & 24.80   & 14.90   & 12.05   & 6.43    & 27.48   & 15.02   & 19.79  & 9.94 \\
         & w/ Context        & 79.81   & 59.97   & 64.60   & 51.12   & 56.10   & 38.67   & 64.64   & 51.38   & 66.29  & 50.29 \\
        \midrule
        $40\times$ & AutoCompressor    & 21.46   & 0.35    & 16.29   & 0.29    & 14.09   & 2.00    & 25.57   & 0.63    & 19.35  & 0.82 \\
        $\textgreater 64\times$ & xRAG              & 18.19   & 3.46    & 27.51   & 16.29   & 13.75   & 3.47    & 38.06   & 20.80   & 24.38  & 11.01 \\
        \midrule
        \multirow{6}{*}{$4\times$} & ICAE (7B)          & 45.69   & 21.63   & 35.16   & 26.68   & 27.98   & 11.70   & 59.15   & 47.35   & 42.00  & 26.84 \\
        & LLMLingua2(0.6B)    & 51.20   & 32.18   & \textbf{55.72}   &\underline{44.18}   & 35.41   & 24.80   & 68.44   & 55.85   & 52.69  & 39.25 \\
         & COCOM-Lite (0.1B)     & 21.70   & 9.17    & 40.07   & 32.32   & 19.45   & 13.90   & 50.45   & 41.87   & 32.92  & 24.32 \\
        & PCC-Lite (0.77B)       & 75.83   & 57.44   & 50.37   & 42.20   & \underline{50.36}   & 37.83   & 70.51   & 61.97   & 61.77  & 49.86 \\
        & PCC-Large (8B)      & \underline{77.76}   & \underline{60.04}   & 48.19   & 39.97   & \textbf{52.56}   & \textbf{39.37}   & \textbf{75.96}   & \textbf{67.71}   & \underline{63.62}  & \underline{51.77} \\
        & OURS (\sys) (0.5B)      & \textbf{79.03}   & \textbf{60.14}   & \underline{54.66}   & \textbf{46.61}   & 50.28   & \underline{38.13}   & \underline{71.82}   & \underline{62.47}   & \textbf{63.95}  & \textbf{51.84} \\
        \midrule
        \multirow{5}{*}{$16\times$} & COCOM-Lite (0.1B)    & 19.23   & 8.13    & 31.94   & 25.27   & 19.45   & 13.90   & 26.36   & 20.66   & 24.22  & 17.20 \\
        & PCC-Lite (0.77B)      & 56.82   & 37.72   & 35.67   & 27.88   & \textbf{40.94}   & \underline{28.07}   & 62.56   & 52.73   & 49.00  & 36.60 \\
        & PCC-Large (8B)     & 55.32   & 37.72   & 33.76   & 25.27   & 38.96   & 26.47   & \textbf{71.09}   & \textbf{61.97}   & 49.78  & 37.86 \\
        & OURS (\sys) (0.5B)     & \textbf{59.41}   & \textbf{39.86}   & \textbf{41.05}   & \textbf{33.24}   & 39.98   & 27.73   & 67.61  & 58.37   & \underline{52.01}  & \underline{39.80} \\
        & OURS (\sys) (1B)     & \underline{58.93}   & \underline{39.69}   & \underline{40.61}   & \underline{32.47}   & \underline{40.53}   & \textbf{29.00}   & \underline{68.17}   & \underline{58.90}   & \textbf{52.06}  & \textbf{40.02} \\
        \midrule
        \multirow{2}{*}{$64\times$} & PCC-Lite (0.77B)      & 37.89   & 22.66   & 20.62   & 14.27   & 27.02   & 16.30   & 36.82   & 29.18   & 30.59  & 20.60 \\
         & OURS (\sys) (0.5B)     & \textbf{40.84}   & \textbf{24.36}   & \textbf{30.54}   & \textbf{23.07}   & \textbf{29.54}   & \textbf{19.40}   & \textbf{57.96}   & \textbf{49.15}   & \textbf{39.72}  & \textbf{29.00} \\
        \bottomrule
    \end{tabular}
    }
    \end{threeparttable}  
    \label{tab.2}
\end{table*}

\subsubsection{In-Context Learning}
Unlike RAG-based QA, which primarily demands precise evidence retrieval and local information extraction, In-Context Learning (ICL) necessitates the inference of latent patterns from demonstration examples and the subsequent application of these patterns to new queries via analogical reasoning. Given that RAG emphasizes local detail while ICL prioritizes global pattern recognition, we evaluate this capability separately.

We assess our method across three diverse datasets: GSM8K (grade school math word problems), SST-2 \cite{socher2013recursive} (sentiment analysis), and WSC \cite{levesque2012winograd} (Winograd Schema Challenge for commonsense reasoning). For the ICL task, we fine-tune the pre-trained \sys~ checkpoint using the GSM8K training set.
Our comparative analysis includes several baselines: zero-shot generation, few-shot ICL with maximum context limits of 150 and 750 tokens (uncompressed), and the PCC baseline utilizing 16x compressed in-context examples. To ensure a fair comparison, the PCC baseline employs the identical compressor architecture, training procedure, and downstream decoder as our \sys~ method.

The results are detailed in Table \ref{tab.3}. On average, \sys~ not only surpasses the PCC baseline but also achieves optimal performance exceeding even the uncompressed few-shot baselines. Notably, our method exhibits a substantial advantage on the WSC dataset, indicating superior capability in retaining commonsense reasoning patterns within the compressed memory.
\begin{table}[h]
    \centering
    \caption{Performance Comparison on Three ICL Datasets (Accuracy \%). "Comp." denotes using compressed memory context. Bold indicates the highest accuracy.}
    \label{tab.3}
    \resizebox{\columnwidth}{!}{
    \begin{tabular}{ccccl}
        \toprule
        Methods & GSM8K & SST-2 & WSC &Average\\
        \midrule
        Zero-shot       & 77.81   & 90.57    & 57.69 & 75.36\\
        ICL 150 Tokens  & 83.56   & \textbf{95.41}    & 50.00 & 76.32\\
        ICL 750 Tokens  & 86.10   & 94.95    & 46.15 & 75.73\\
        \midrule
        PCC-16x Comp. 150 Tokens   & 90.11   & 92.09    & 43.27 & 75.16\\
        PCC-16x Comp. 750 Tokens   & \textbf{91.44}   & 90.02    & 47.12 & 76.19\\
        \midrule
        OURS (16x Comp. 150 Tokens) & 89.84   & 88.65    & 57.69 & 78.73\\
        OURS (16x Comp. 750 Tokens) & 90.64   & 87.61    & \textbf{65.38} & \textbf{81.21}\\
        \bottomrule
    \end{tabular}
    }
\end{table}

\subsection{Block-wise Causal Mask Accelerates Compressor Learning}

\subsubsection{Pre-training Dynamics}
\begin{figure}
    \centering
    \includegraphics[width=0.9\linewidth]{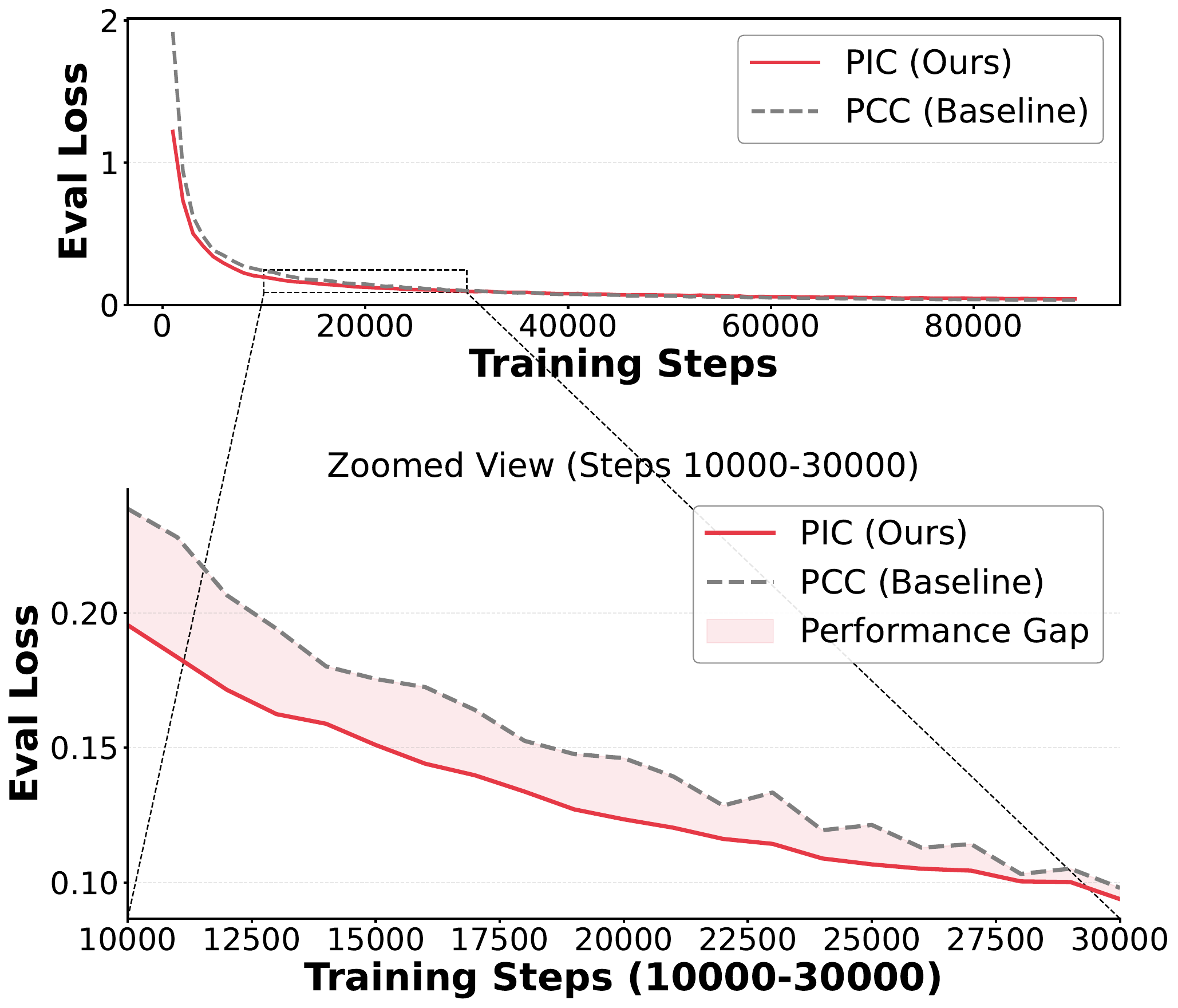}
    \caption{Comparison of Reconstruction Loss During Pre-training. The upper plot illustrates the overall loss trajectory, while the lower plot presents a zoomed-in view of steps 10k to 30k, highlighting the faster convergence of our \sys~ (red) compared to the PCC baseline (dashed gray).}
    \label{fig.5}
\end{figure}

We compare the pre-training loss trajectories to evaluate the learning efficiency of the compressor.
As shown in Figure \ref{fig.5}, our method exhibits a noticeably faster convergence rate during the initial phase of pre-training, achieving an acceleration of approximately $1.1 \times$ to $1.3 \times$ compared to the baseline. This empirically demonstrates that our approach, by explicitly constraining the receptive field to sequential blocks, effectively reduces the optimization complexity and facilitates the learning process for the compressor.
\subsubsection{Data Efficiency Analysis}
As observed in Figure \ref{fig.5}, after 30k training steps, both \sys~ and the PCC baseline approach  convergence, rendering it challenging to discern capability differences based solely on pre-training reconstruction loss. Consequently, to thoroughly investigate data efficiency, we perform domain-specific fine-tuning on checkpoints saved at various pre-training intervals and evaluate their performance on downstream QA tasks. To ensure a fair comparison with PCC, we utilize the identical Qwen2.5-0.5B-Instruct backbone for both methods, isolating the attention mechanism as the sole independent variable.

\begin{figure}
    \centering
    \includegraphics[width=0.8\linewidth]{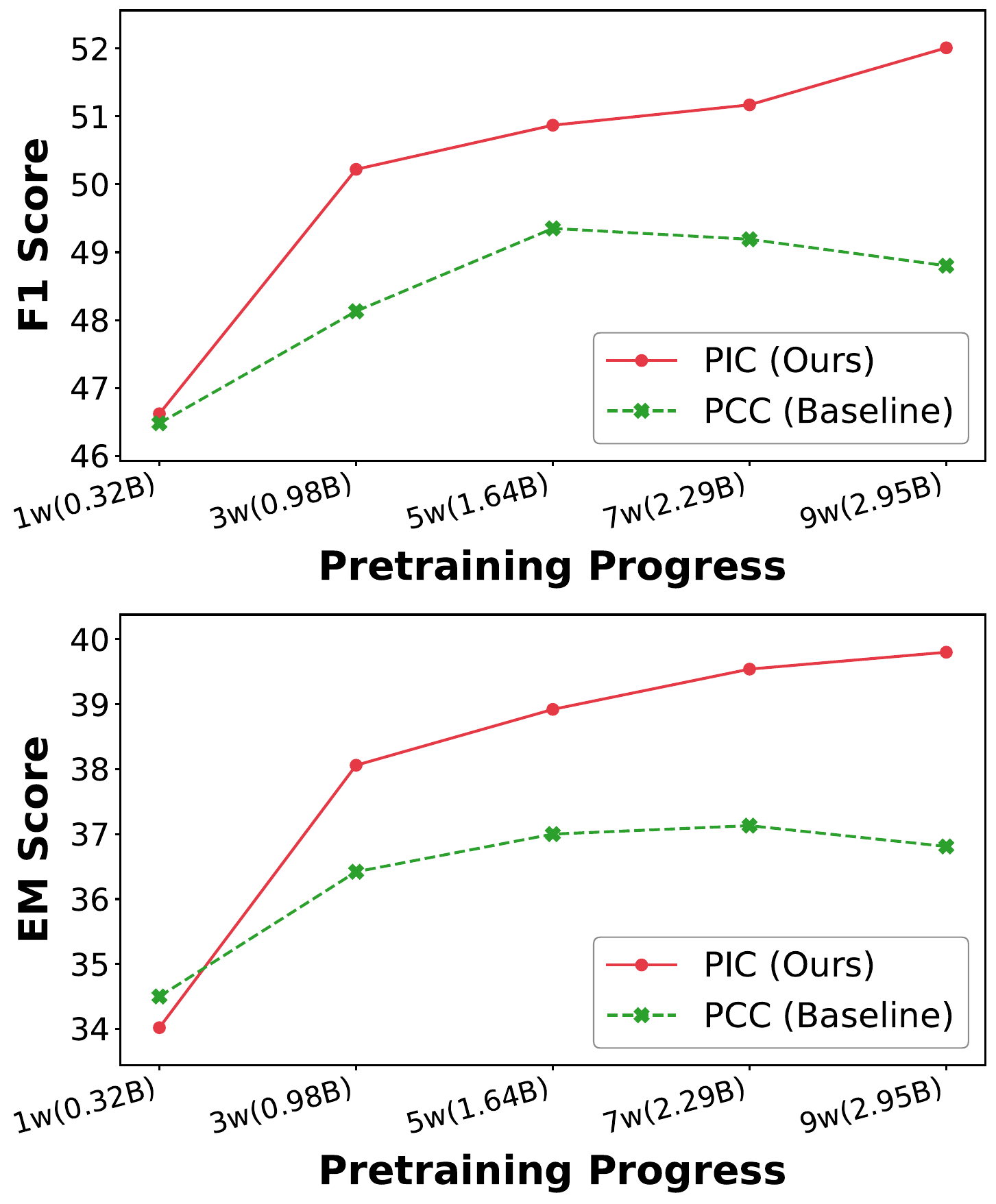}
    \caption{Relationship between Compressor Performance on Downstream QA Tasks and Pre-training Steps. The x-axis represents the number of tokens corresponding to the pre-training steps (converted to 1 epoch equivalent), and the y-axis represents the EM or F1 score.}
    \label{fig.6}
\end{figure}

As demonstrated in Figure \ref{fig.6}, our method achieves superior downstream performance compared to the PCC baseline using less than 1 billion pre-training tokens (Detailed results for other compression rates (4x and 64x) are shown in Figure \ref{fig.a.4x.64x} in Appendix D.). Remarkably, under the hardware setup of two 48GB RTX 4090 GPUs, our compressor requires only 56 hours of training to surpass the peak performance of the baseline, equivalent to a reduction in training time of approximately 40\%. Furthermore, the performance of our method exhibits a consistent upward trend as the amount of pre-training data increases. In contrast, the PCC baseline fails to maintain a stable improvement with increased data, even showing performance degradation after 50k steps. These results collectively validate that our method significantly enhances both the data efficiency and training stability of the compressor.

\subsection{Analysis of Learned Embeddings}
\begin{figure}
    \centering
    \includegraphics[width=1\linewidth]{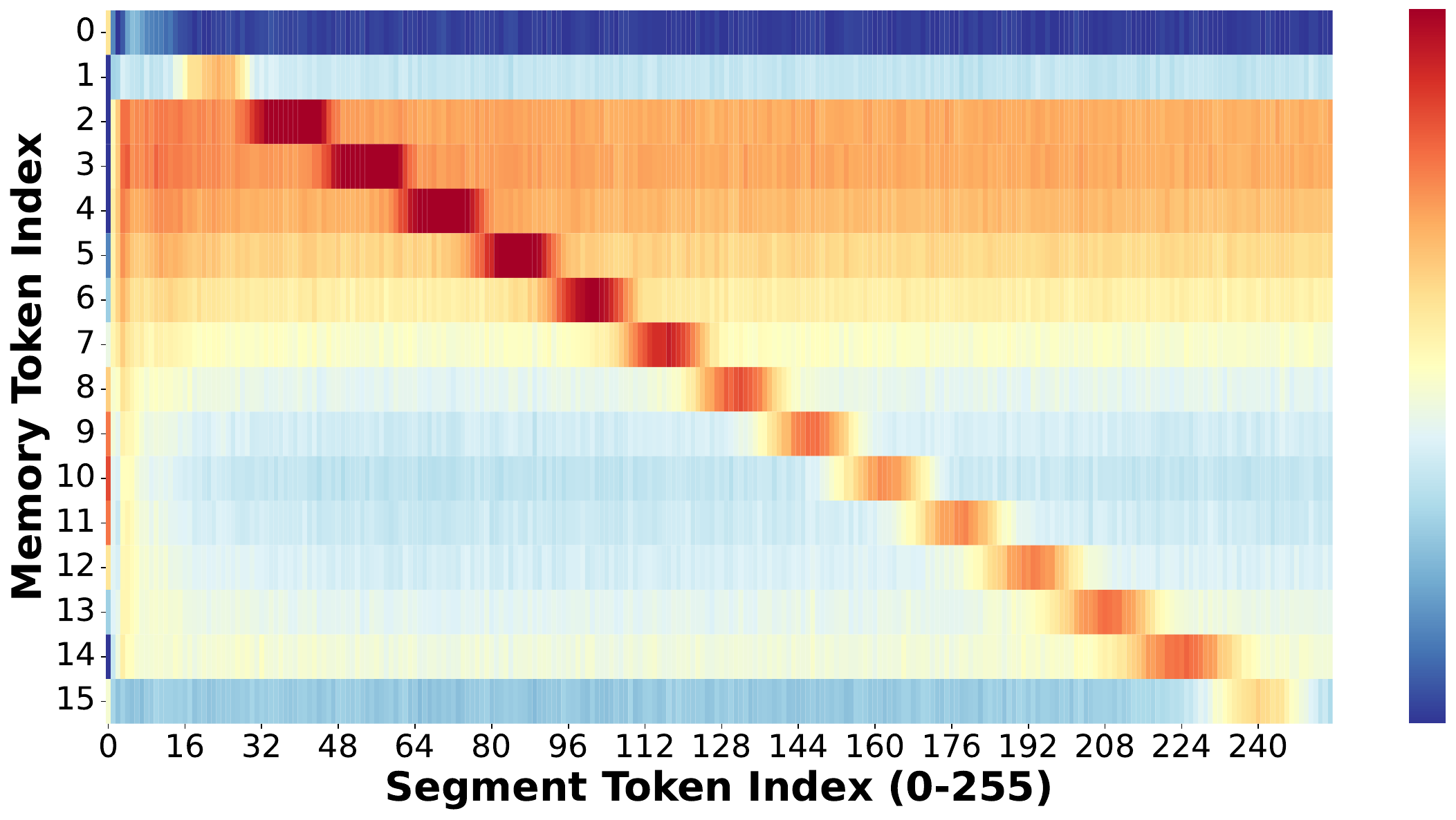}
    \caption{Heatmap visualization of the cosine similarity between memory embeddings (\sys) generated by the Fully Trained Compressor (90k Steps) and original token embeddings on the HotpotQA dataset. Red indicates higher similarity, while blue represents lower similarity.}
    \label{fig.7}
\end{figure}

Different from the claim by Louis et al.~\cite{louis-etal-2025-pisco} that spatial specialization does not occur when the decoder is frozen, we hypothesize that such spatial specialization is, in fact, a critical indicator of effective compression.
We calculated the cosine similarity between the memory embeddings generated by \sys~ and the original context token embeddings (For additional visualization results of spatial specialization across other datasets, please refer to Appendix \ref{app:pic}). As visualized in Figure \ref{fig.7}, the learned attention patterns align perfectly with our imposed receptive field constraints: each memory token sequentially focuses intensively on the specific text chunk it is designated to compress.

\begin{figure}
    \centering
    \captionsetup[subfigure]{skip=2pt} 
    \captionsetup[subfigure]{font=small}
    \begin{subfigure}{1\linewidth}
        \centering
        \includegraphics[width=\textwidth]{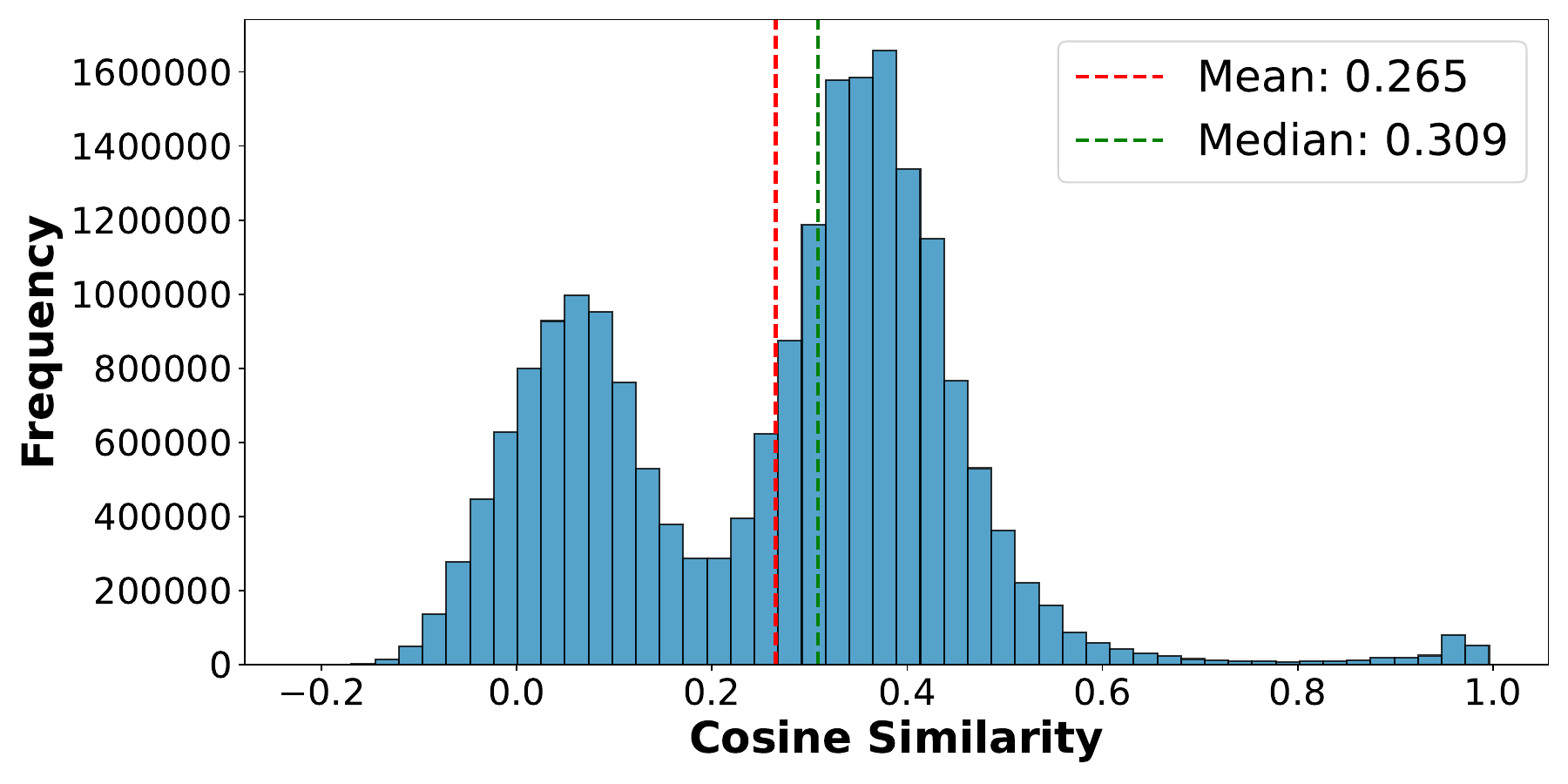}
        \caption{\textbf{PCC}}
    \end{subfigure}
    
    \vspace{2pt} 
    
    \begin{subfigure}{1\linewidth}
        \centering
        \includegraphics[width=\textwidth]{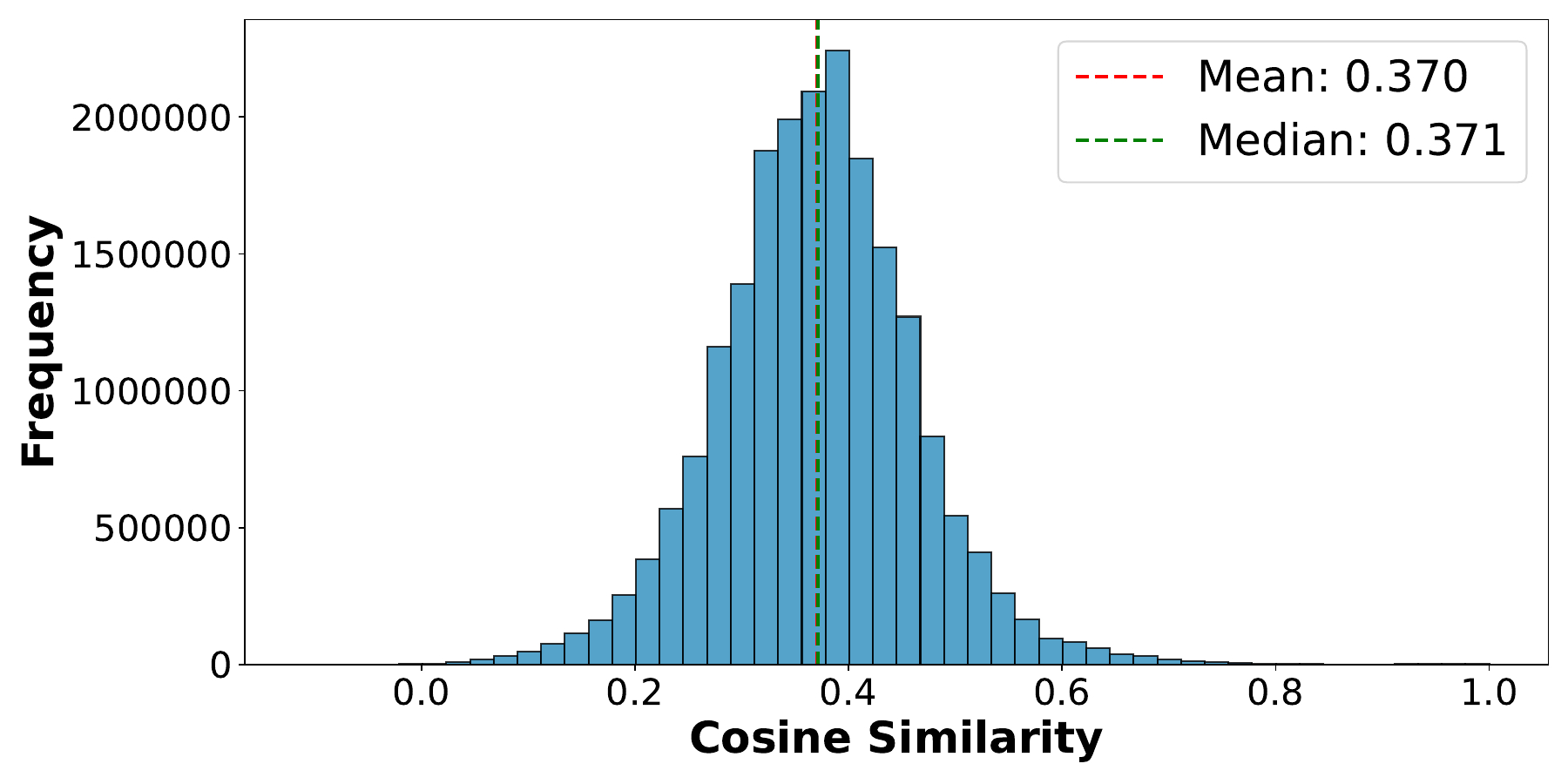}
        \caption{\textbf{OURS (\sys)}}
    \end{subfigure}
    
    \setlength{\abovecaptionskip}{5pt} 
    \caption{Distribution of pairwise cosine similarities between memory embeddings within the same memory slot on the HotpotQA dataset. Compressors are fine-tuned on downstream QA tasks.}
    \label{fig.8}
\end{figure}

Furthermore, we investigated the internal coherence of the memory slots by analyzing the pairwise cosine similarity between memory tokens $(m_i, m_j)$ (We also conducted this analysis on other datasets; for detailed results, please refer to Appendix \ref{app:m2m}). As illustrated in Figure \ref{fig.8}, the memory embeddings generated by the PCC exhibit a problematic distribution. 
Specifically, The PCC distribution displays a heavy right tail, indicating high similarity and redundancy among certain memory embeddings.
Conversely, the presence of negative cosine similarities on the left side suggests potential adversarial semantics between some memory tokens.
Most notably, the distribution exhibits a bimodal characteristic; the left peak, most memory embeddings have similarities between 0 and 0.2, implying that a portion of the memory embeddings may serve as redundant noise.

In contrast, our \sys~ method yields a distribution that approximates a normal distribution, significantly mitigating potential adversarial semantics, low-similarity noise and the high-similarity redundancy. This statistical evidence suggests that our block-wise approach facilitates the learning of more distinct, orthogonal, and information-rich representations for each memory token.

\subsection{Ablation Study}
To strictly isolate the effectiveness of our proposed block-wise causal mask, we conducted an ablation study at a 16$\times$ compression ratio. We utilized the representative Qwen2.5-0.5B-Instruct backbone for all variants to factor out potential performance gaps caused by different model architectures, strictly varying only the attention mask configuration.
The baseline configuration, denoted as Full Causal Mask, retains the standard global causal attention mechanism utilized in the original PCC framework.

As shown in Table \ref{tab.4} (for the complete table, please refer to Table \ref{tab.a.1} in the Appendix), our method significantly outperforms the Full Causal Mask baseline. This comparison confirms that the observed performance gains are directly due to our proposed masking strategy rather than the choice of the compressor backbone.
\begin{table}
    \centering
    \caption{Ablation Study: Average performance across four datasets in the RAG-based QA domain. Reported metrics are in percentage (\%).}
    \label{tab.4}
    \begin{tabular*}{\linewidth}{@{\extracolsep{\fill}}ccl}
        \toprule
        Method & F1 & EM \\
        \midrule
        16x-\sys & \textbf{52.01} & \textbf{39.80} \\
        Full Causal Mask (16x-PCC) & 48.80 & 36.81 \\
        \bottomrule
    \end{tabular*}
\end{table}

\section{Conclusion}
Drawing inspiration from the chunking mechanism in human working memory and the observed spatial specialization of memory embeddings relative to original tokens, we introduced Parallelized Iterative Compression (\sys). By implementing a straightforward modification to the Transformer attention mask, \sys~ explicitly constrains the receptive field of memory tokens to sequential local blocks. This structural inductive bias significantly reduces the optimization complexity inherent in compressor training.
Empirical results across multiple downstream tasks demonstrate that our method outperforms a wide range of competitive baselines. This advantage is particularly pronounced in high-compression scenarios; for example, with a compression ratio of 64$\times$, \sys~ achieves relative improvements of 29.8\% in F1 score and 40.7\% in EM score on QA tasks. Furthermore, our approach substantially accelerates the training process. Specifically, for the 16$\times$ compressor, our method surpasses the baseline's peak performance with approximately 40\% less training time, demonstrating significantly improved data efficiency. We also observed that extending training with additional data yields continuous performance gains. Finally, we shed light on the intrinsic properties of text embeddings, suggesting that the phenomenon of spatial specialization may serve as a critical indicator of compressor efficacy. We hope these findings will offer valuable insights to inspire future research in the field of context compression.


\bibliographystyle{ACM-Reference-Format}
\bibliography{reference}

\appendix

\section{Heatmap Visualization of Spatial Specialization Phenomena in ICAE}
\begin{figure*}
    \centering
    \includegraphics[width=0.8\linewidth]{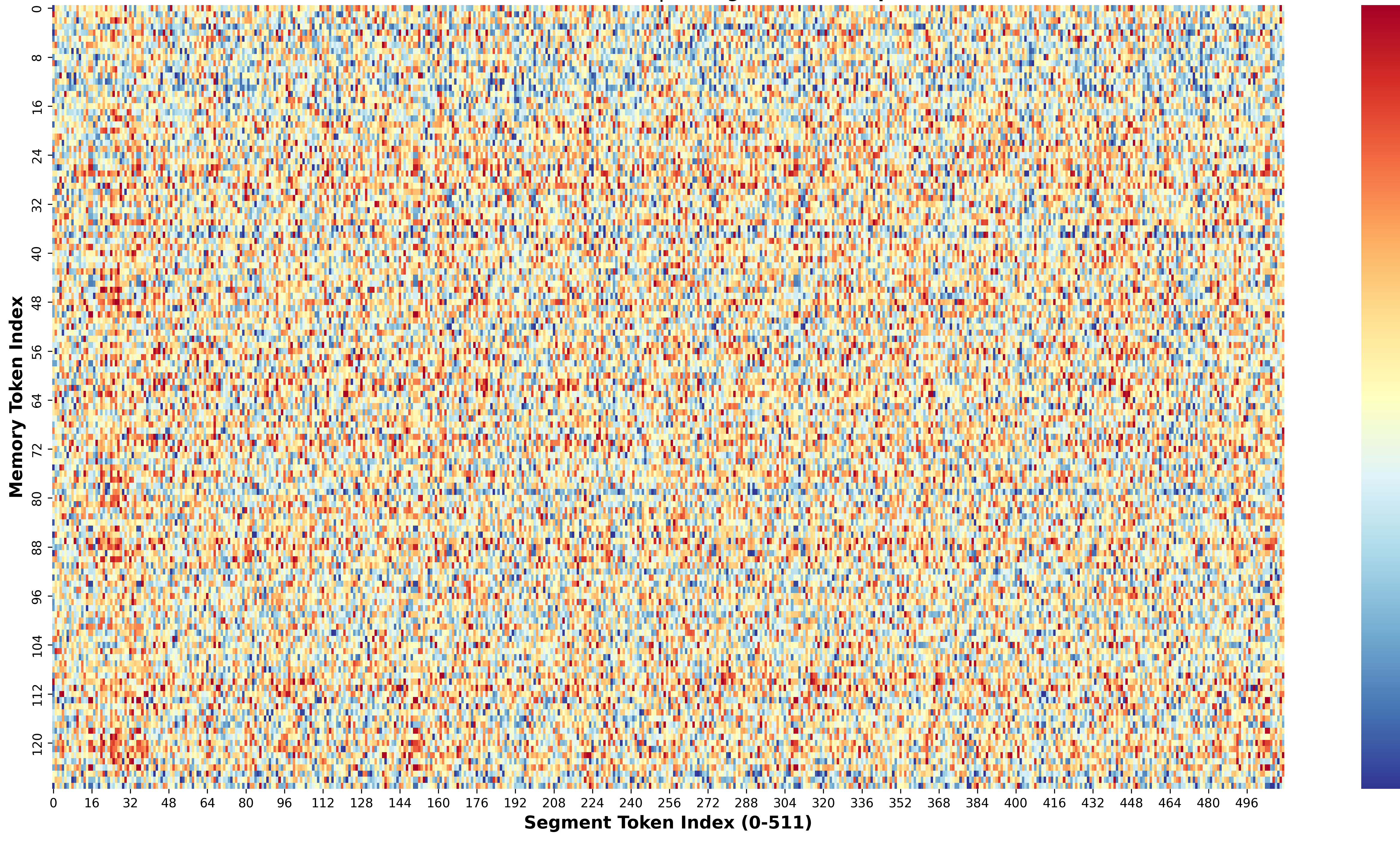}
    \caption{Heatmap visualization of the cosine similarity between memory embeddings (PCC) and original token embeddings on the SQuAD dataset. Red indicates higher similarity, while blue represents lower similarity.}
    \label{fig.icae}
\end{figure*}

We also conducted a cosine similarity analysis between memory embeddings and original token embeddings on ICAE.The number of memory tokens in the fixed memory slots of ICAE is set to 128. We analyzed texts with a length of 512. As shown in Figure \ref{fig.icae}, the image appears chaotic, and no obvious spatial specialization phenomenon was observed.

\section{Heatmap Visualization of Spatial Specialization Phenomena on Other Datasets}

\subsection{PCC}\label{app:pcc}
\begin{figure*}
    \centering
    \captionsetup[subfigure]{font=normalsize}

    \begin{subfigure}{0.4\linewidth}
        \centering

        \includegraphics[width=\textwidth]{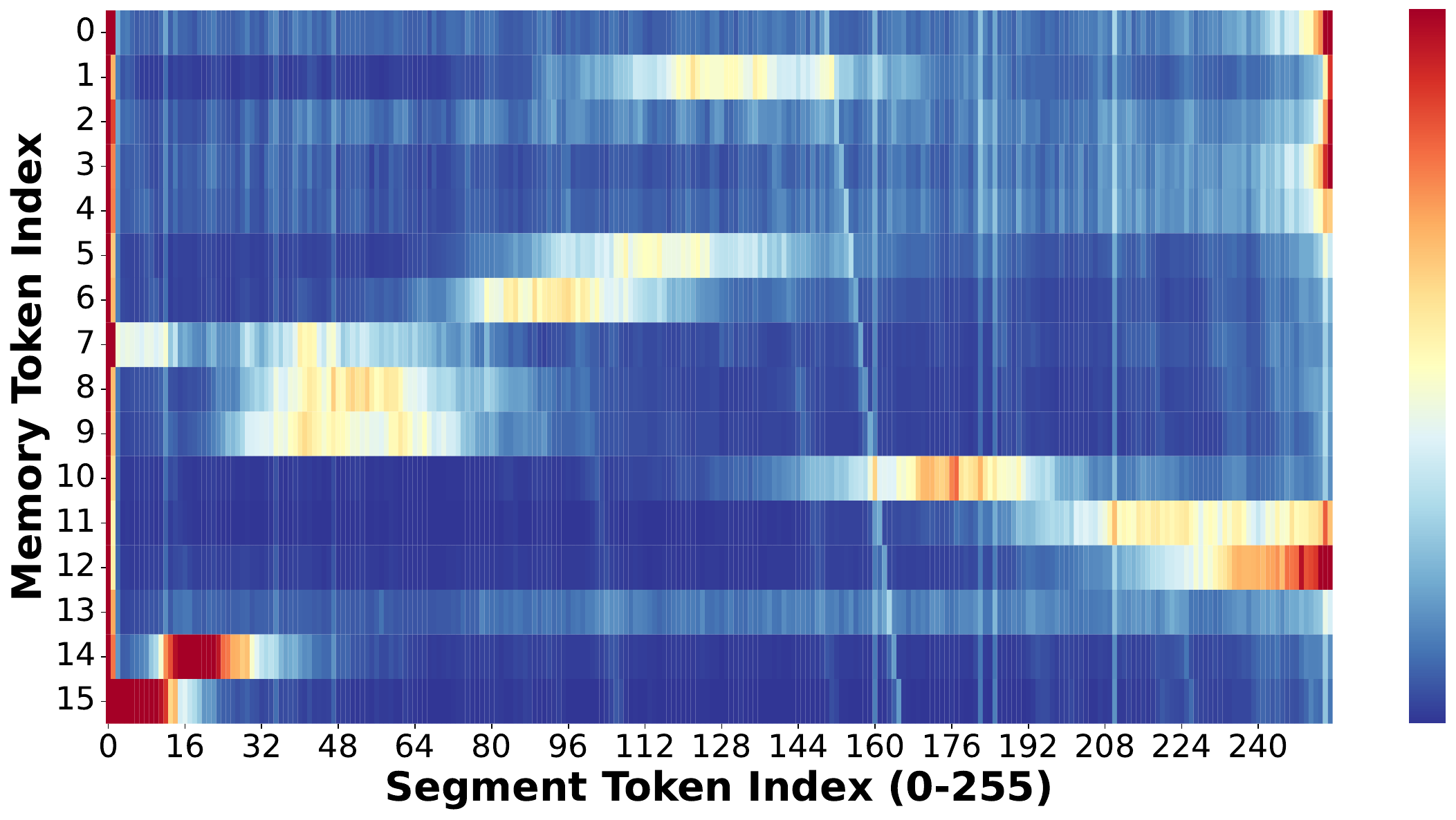}

        \caption{\textbf{pre-trained compressor}}
    \end{subfigure}
    
    \vspace{2pt} 
    
    \begin{subfigure}{0.4\linewidth}
        \centering
        \includegraphics[width=\textwidth]{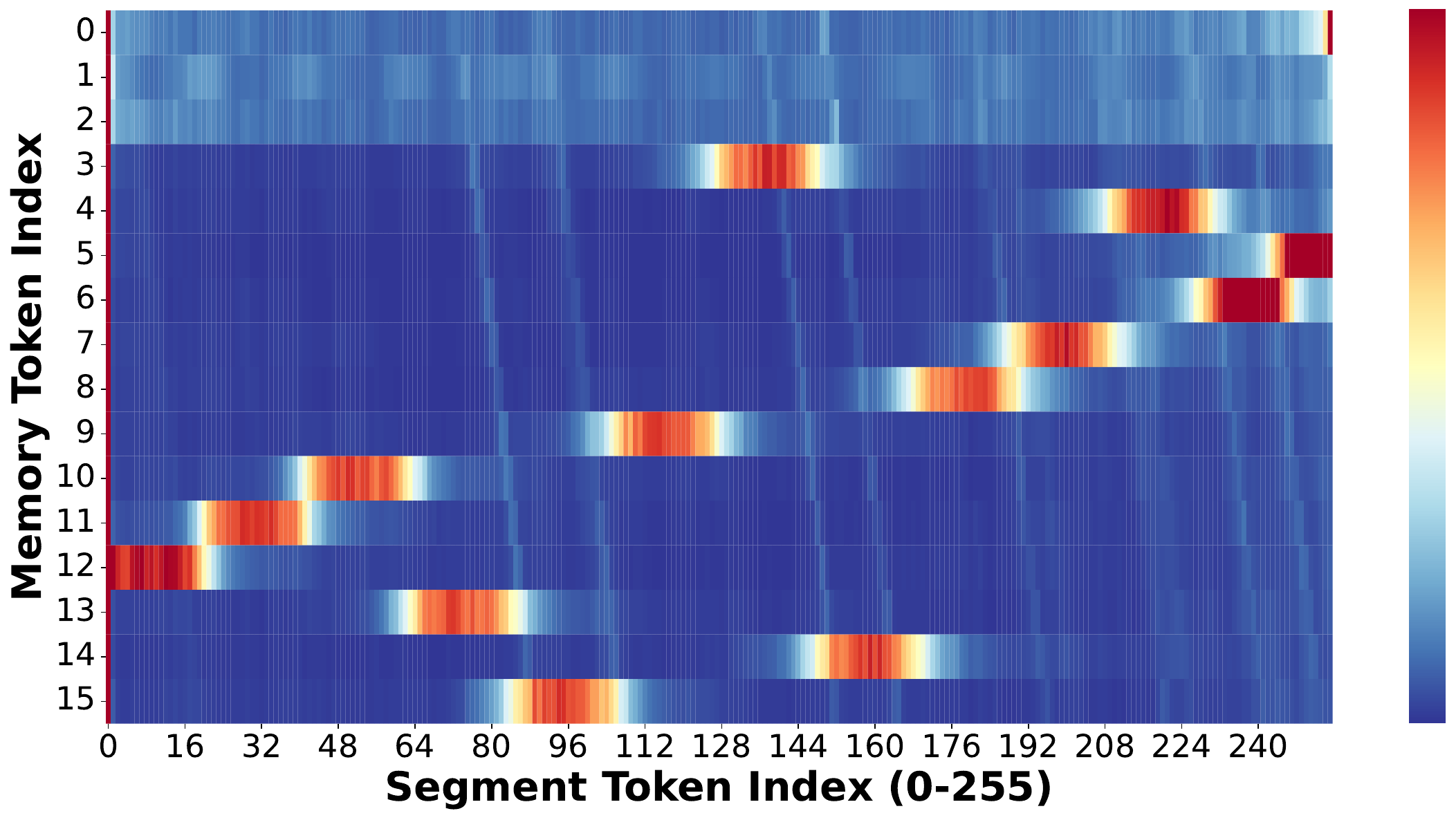}
        \caption{\textbf{fine-tuned compressor}}
    \end{subfigure}
    
    \caption{Attention weight heatmap between memory tokens and original context tokens on the SQuAD dataset.}

    \label{fig.a.pcc.attn.s}
\end{figure*}

\begin{figure*}
    \centering
    \captionsetup[subfigure]{font=normalsize}

    \begin{subfigure}{0.4\linewidth}
        \centering

        \includegraphics[width=\textwidth]{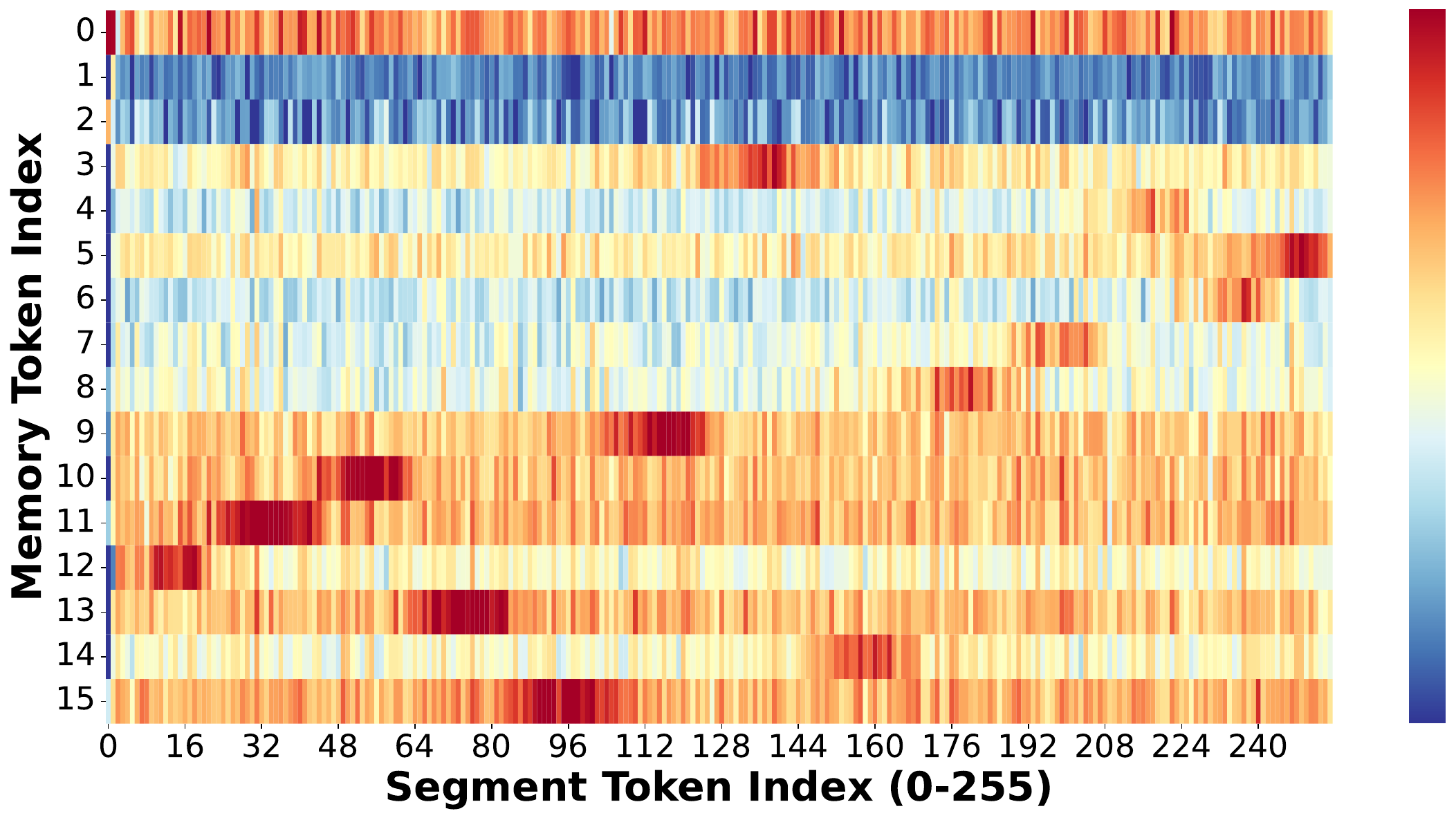}

        \caption{\textbf{pre-trained compressor}}
    \end{subfigure}
    
    \vspace{2pt} 
    
    \begin{subfigure}{0.4\linewidth}
        \centering
        \includegraphics[width=\textwidth]{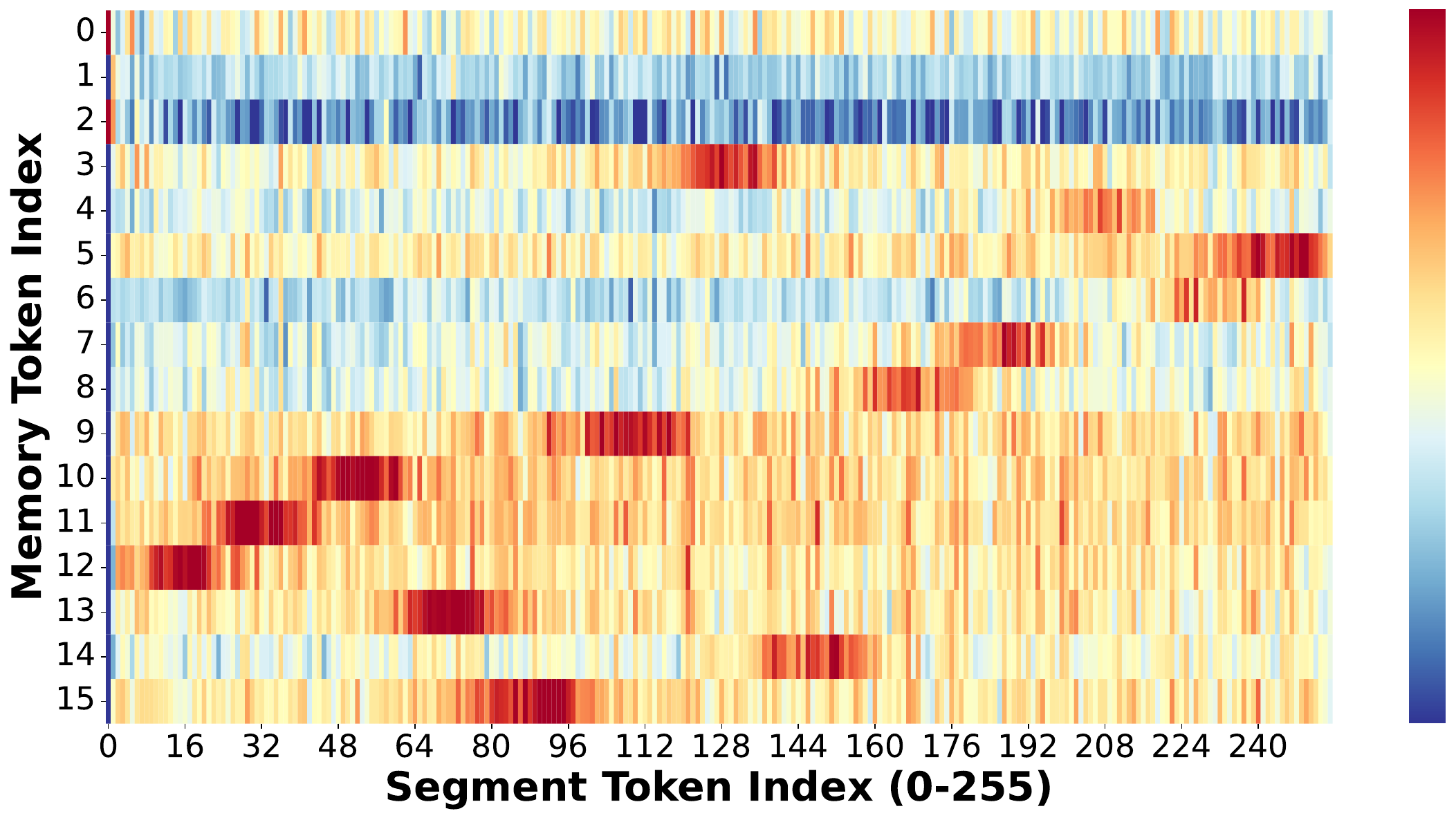}
        \caption{\textbf{fine-tuned compressor}}
    \end{subfigure}
    
    \caption{Heatmap visualization of the cosine similarity between memory embeddings (PCC) and original token embeddings on the SQuAD dataset. Red indicates higher similarity, while blue represents lower similarity.}

    \label{fig.a.pcc.s}
\end{figure*}

\begin{figure*}
    \centering
    \captionsetup[subfigure]{font=normalsize}

    \begin{subfigure}{0.4\linewidth}
        \centering

        \includegraphics[width=\textwidth]{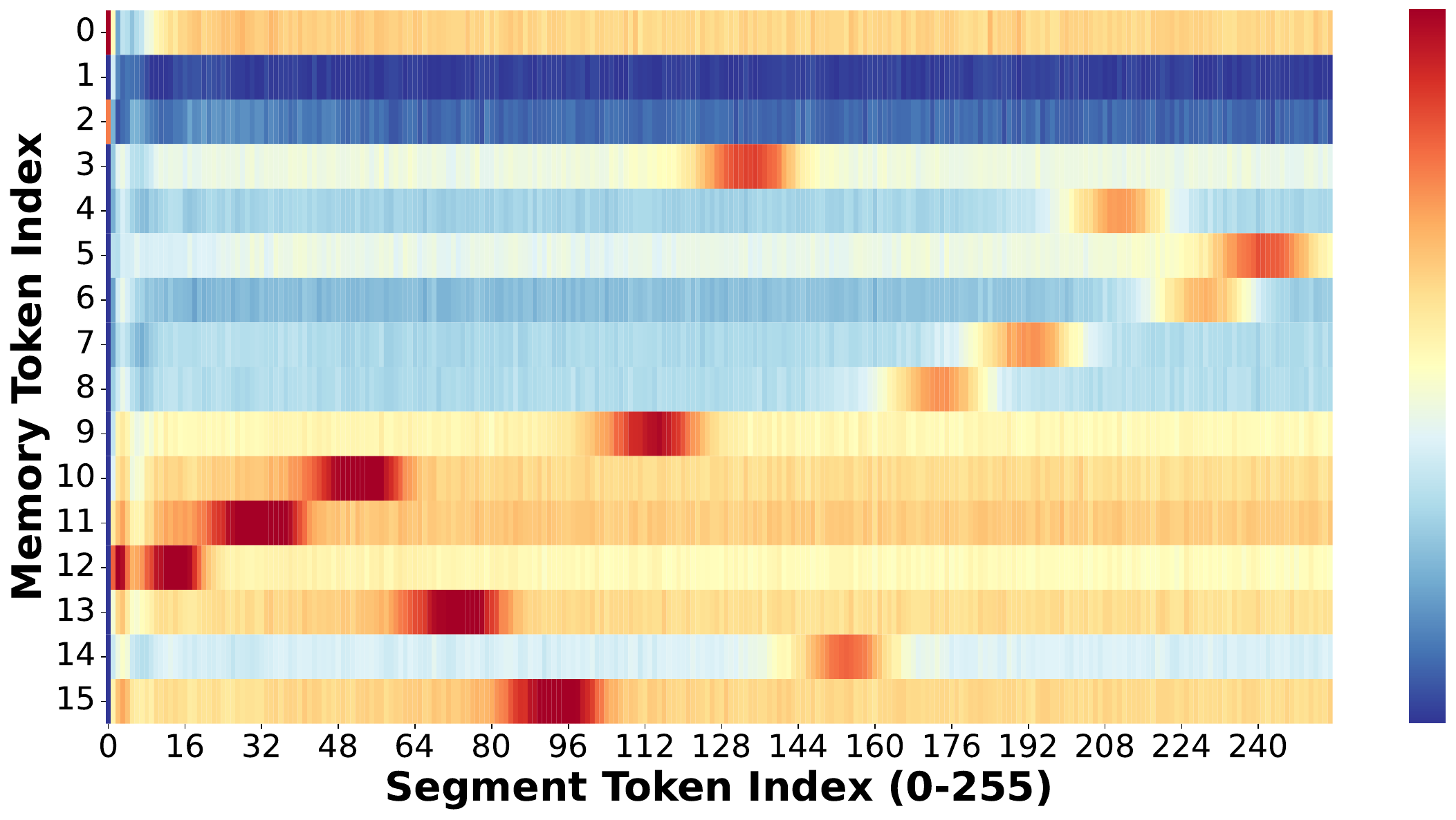}

        \caption{\textbf{pre-trained compressor}}
    \end{subfigure}
    
    \vspace{2pt} 
    
    \begin{subfigure}{0.4\linewidth}
        \centering
        \includegraphics[width=\textwidth]{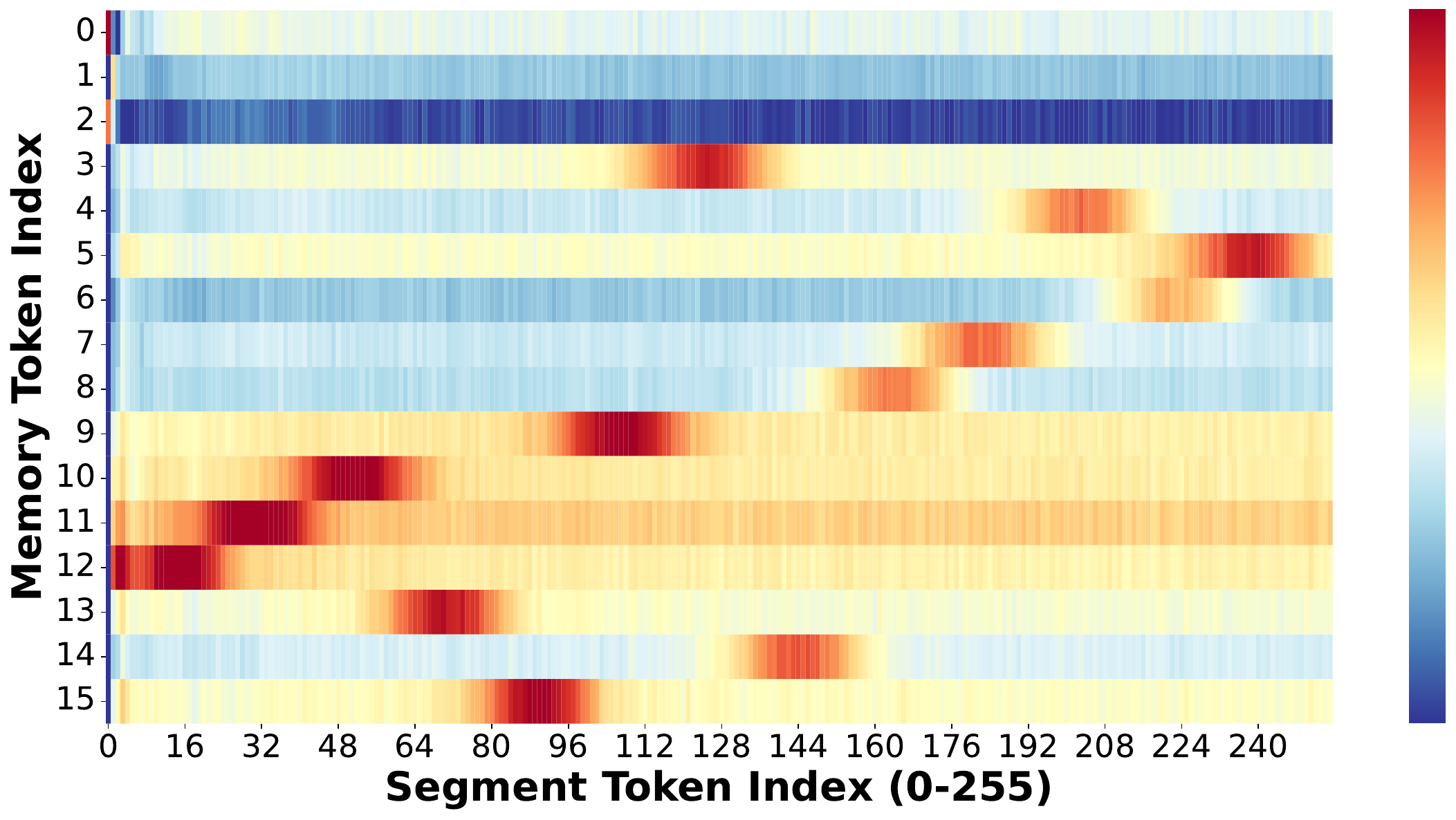}
        \caption{\textbf{fine-tuned compressor}}
    \end{subfigure}
    
    \caption{Heatmap visualization of the cosine similarity between memory embeddings (PCC) and original token embeddings on the HotpotQA dataset. Red indicates higher similarity, while blue represents lower similarity.}

    \label{fig.a.pcc.h}
\end{figure*}

\begin{figure*}
    \centering
    \captionsetup[subfigure]{font=normalsize}

    \begin{subfigure}{0.4\linewidth}
        \centering

        \includegraphics[width=\textwidth]{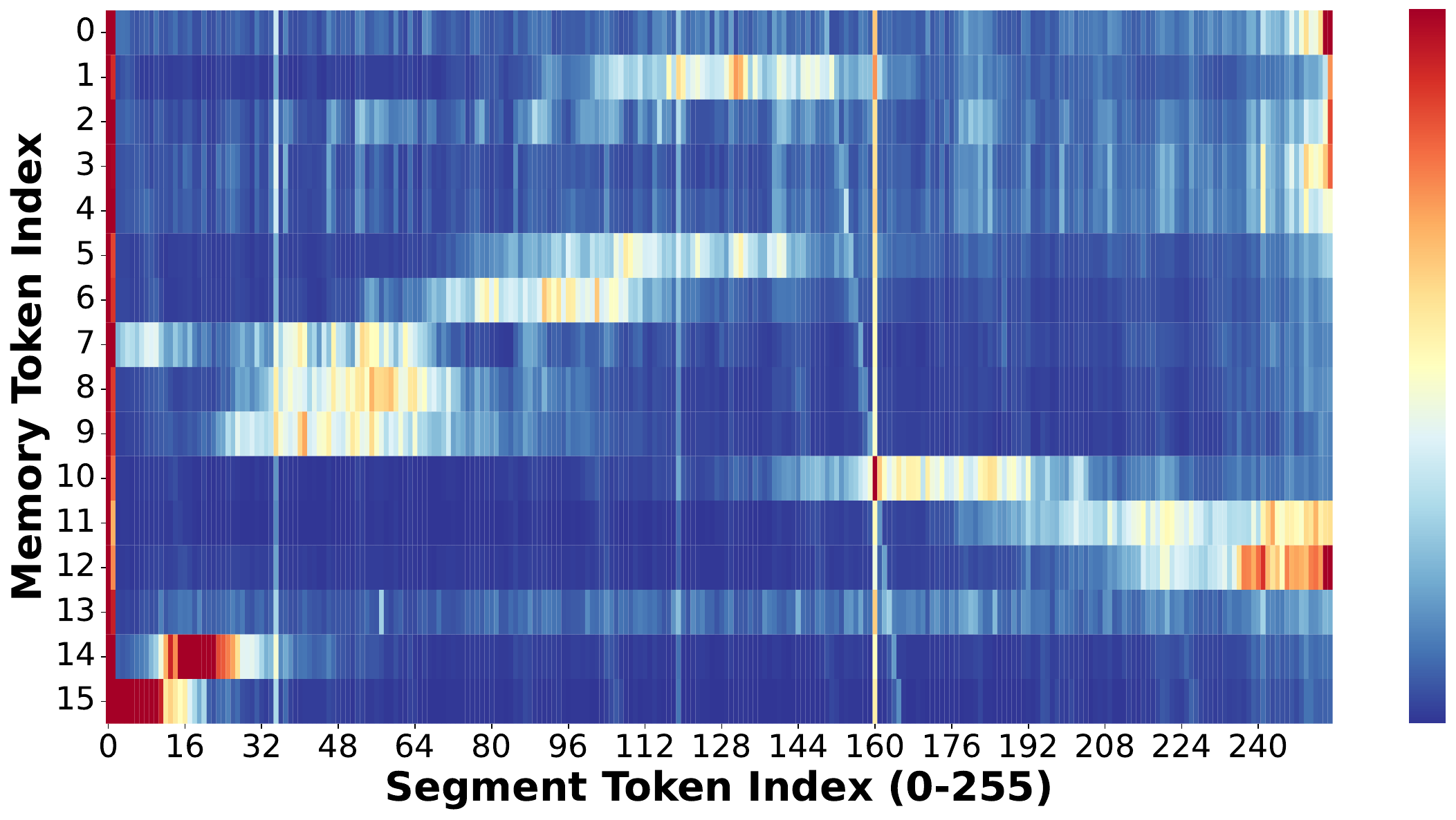}

        \caption{\textbf{pre-trained compressor}}
    \end{subfigure}
    
    \vspace{2pt} 
    
    \begin{subfigure}{0.4\linewidth}
        \centering
        \includegraphics[width=\textwidth]{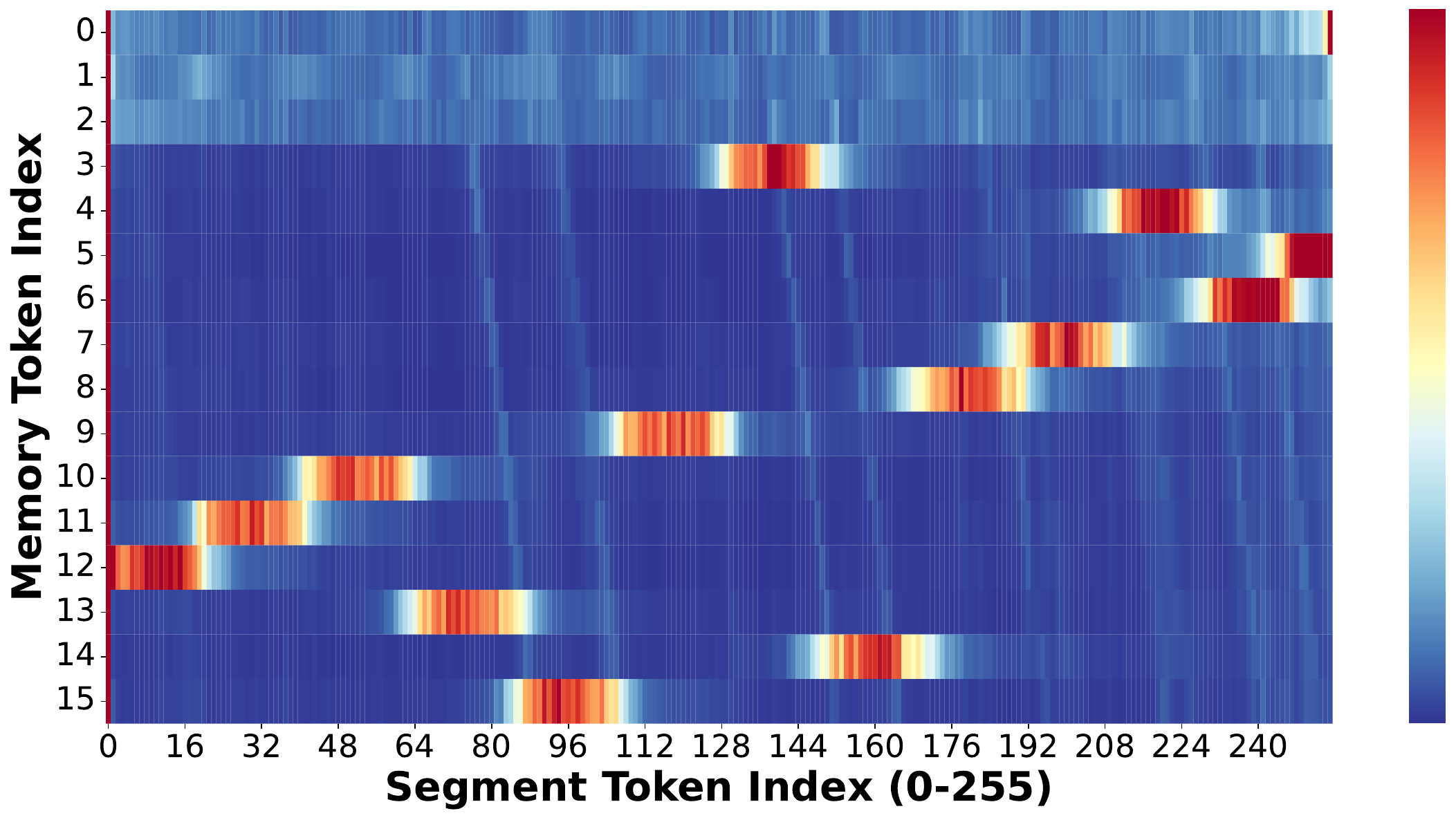}
        \caption{\textbf{fine-tuned compressor}}
    \end{subfigure}
    
    \caption{Attention weight heatmap between memory tokens and original context tokens on the AdversarialQA dataset.}

    \label{fig.a.pcc.attn.a}
\end{figure*}

\begin{figure*}
    \centering
    \captionsetup[subfigure]{font=normalsize}

    \begin{subfigure}{0.4\linewidth}
        \centering

        \includegraphics[width=\textwidth]{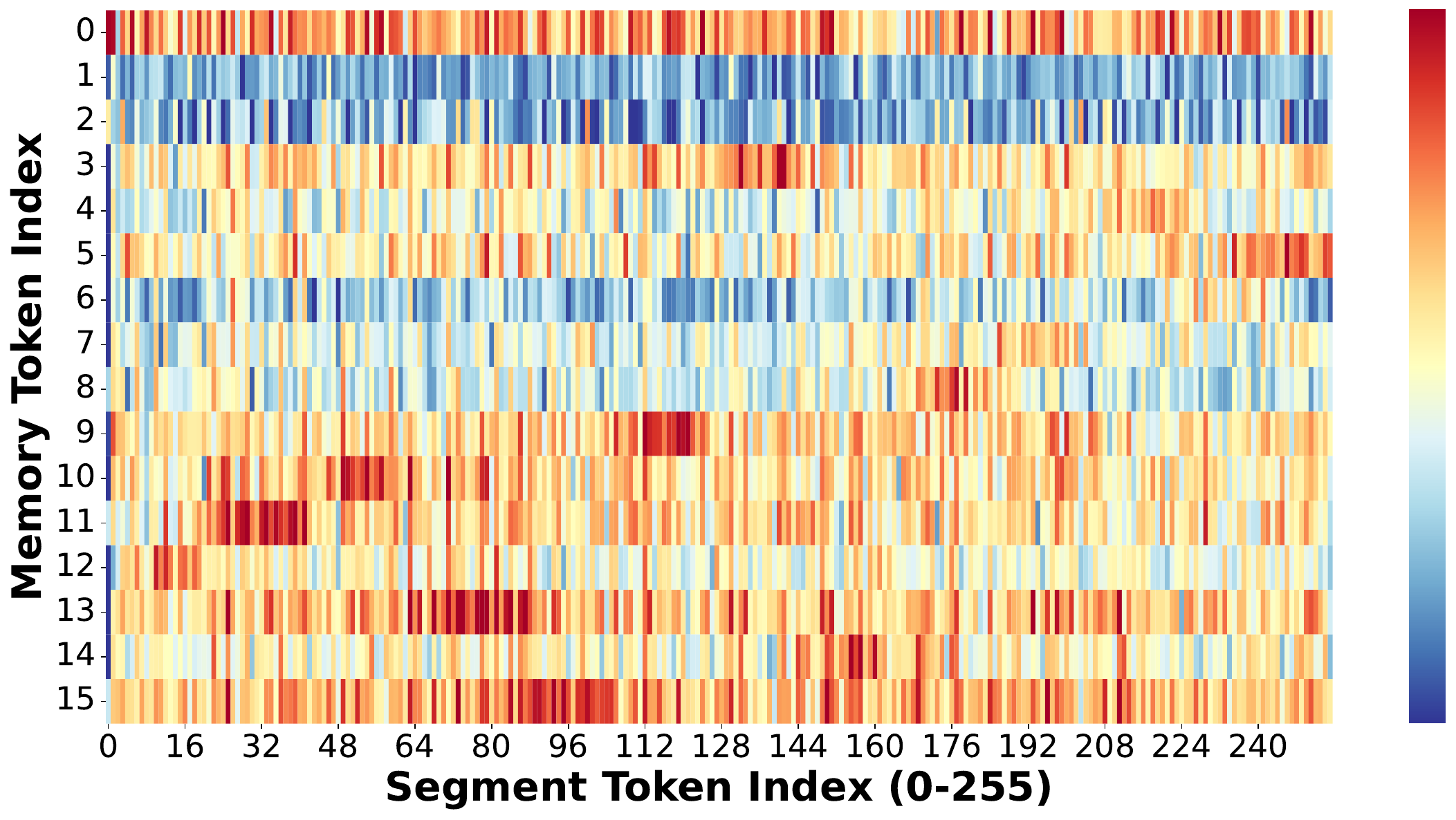}

        \caption{\textbf{pre-trained compressor}}
    \end{subfigure}
    
    \vspace{2pt} 
    
    \begin{subfigure}{0.4\linewidth}
        \centering
        \includegraphics[width=\textwidth]{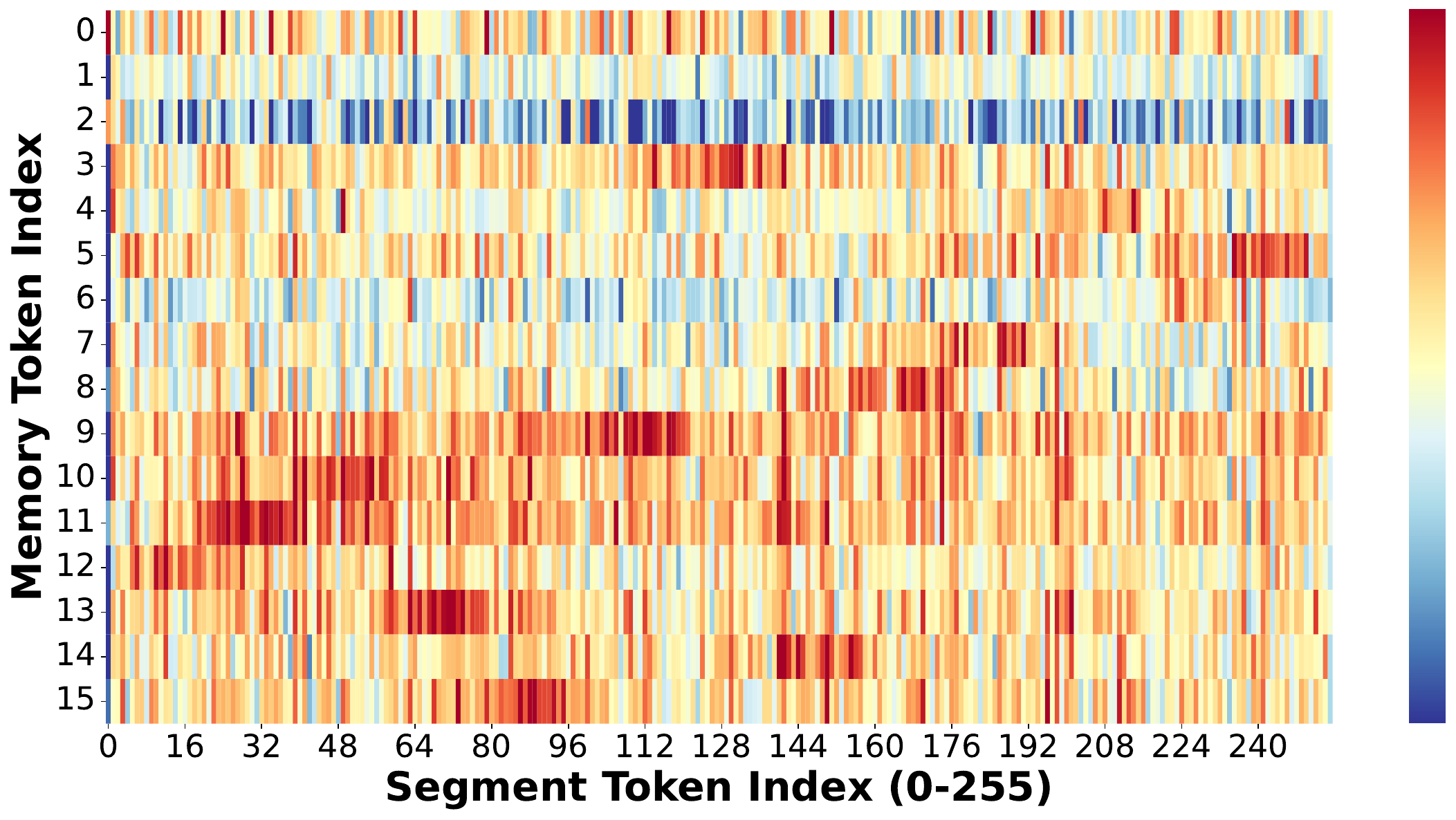}
        \caption{\textbf{fine-tuned compressor}}
    \end{subfigure}
    
    \caption{Heatmap visualization of the cosine similarity between memory embeddings (PCC) and original token embeddings on the AdversarialQA dataset. Red indicates higher similarity, while blue represents lower similarity.}
    \label{fig.a.pcc.a}
\end{figure*}

\begin{figure*}[h]
    \centering

    \captionsetup[subfigure]{font=normalsize}

    \begin{subfigure}{0.4\linewidth}
        \centering

        \includegraphics[width=\textwidth]{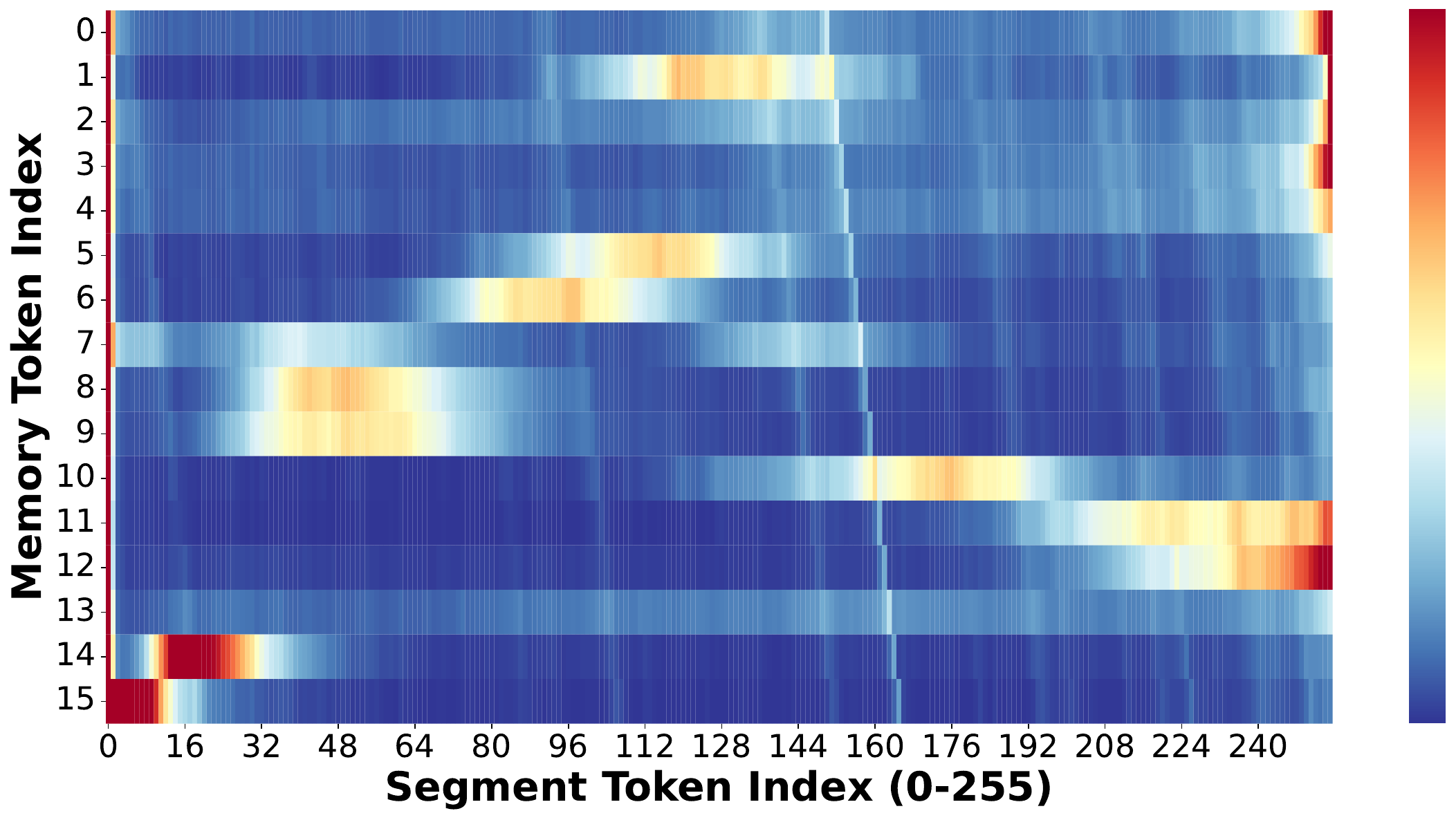}

        \caption{\textbf{pre-trained compressor}}
    \end{subfigure}
    
    \vspace{2pt} 
    
    \begin{subfigure}{0.4\linewidth}
        \centering
        \includegraphics[width=\textwidth]{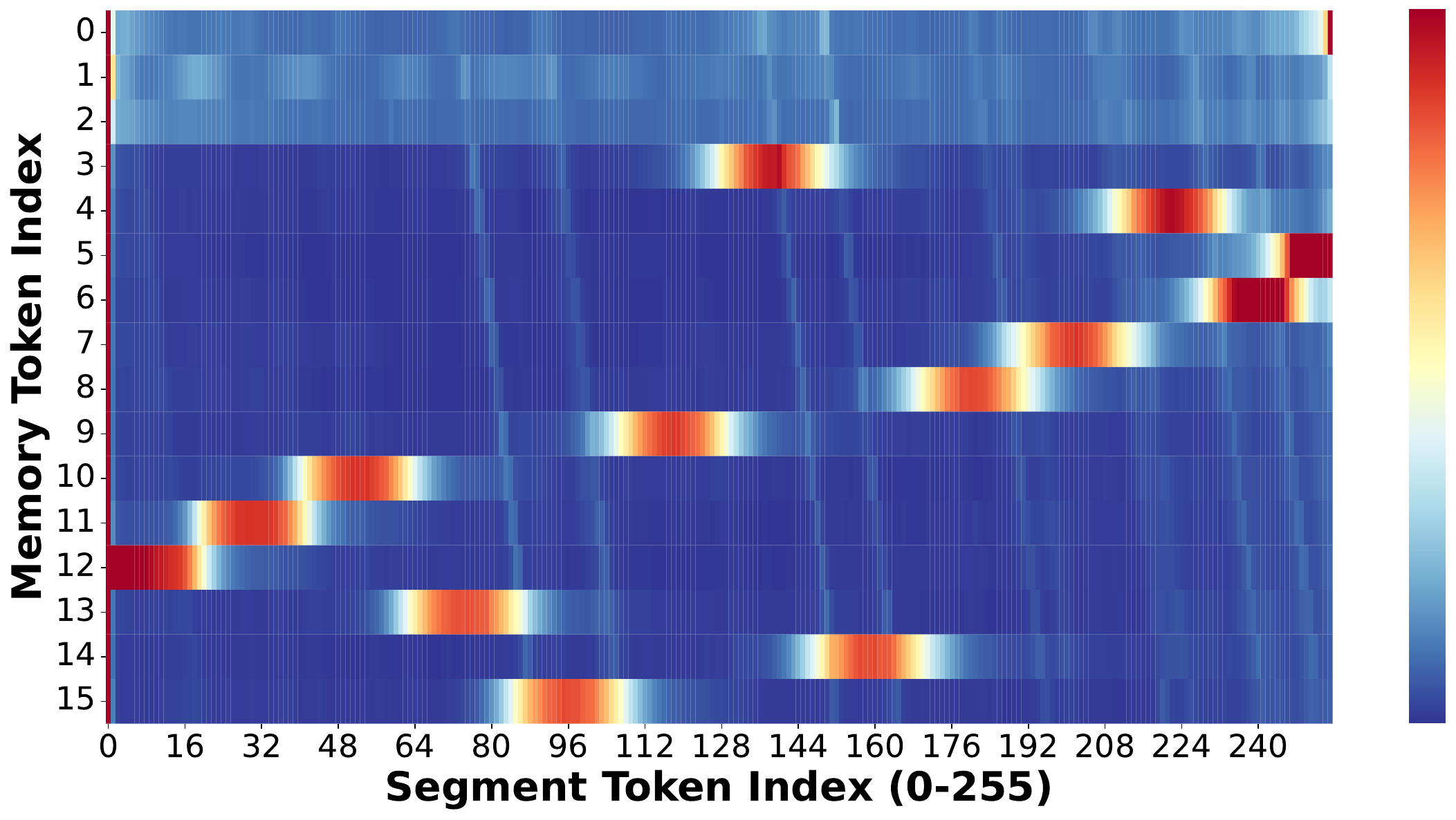}
        \caption{\textbf{fine-tuned compressor}}
    \end{subfigure}
    
    \caption{Attention weight heatmap between memory tokens and original context tokens on the Natural Questions dataset.}
    \label{fig.a.pcc.attn.n}
\end{figure*}

\begin{figure*}
    \centering

    \captionsetup[subfigure]{font=normalsize}

    \begin{subfigure}{0.4\linewidth}
        \centering

        \includegraphics[width=\textwidth]{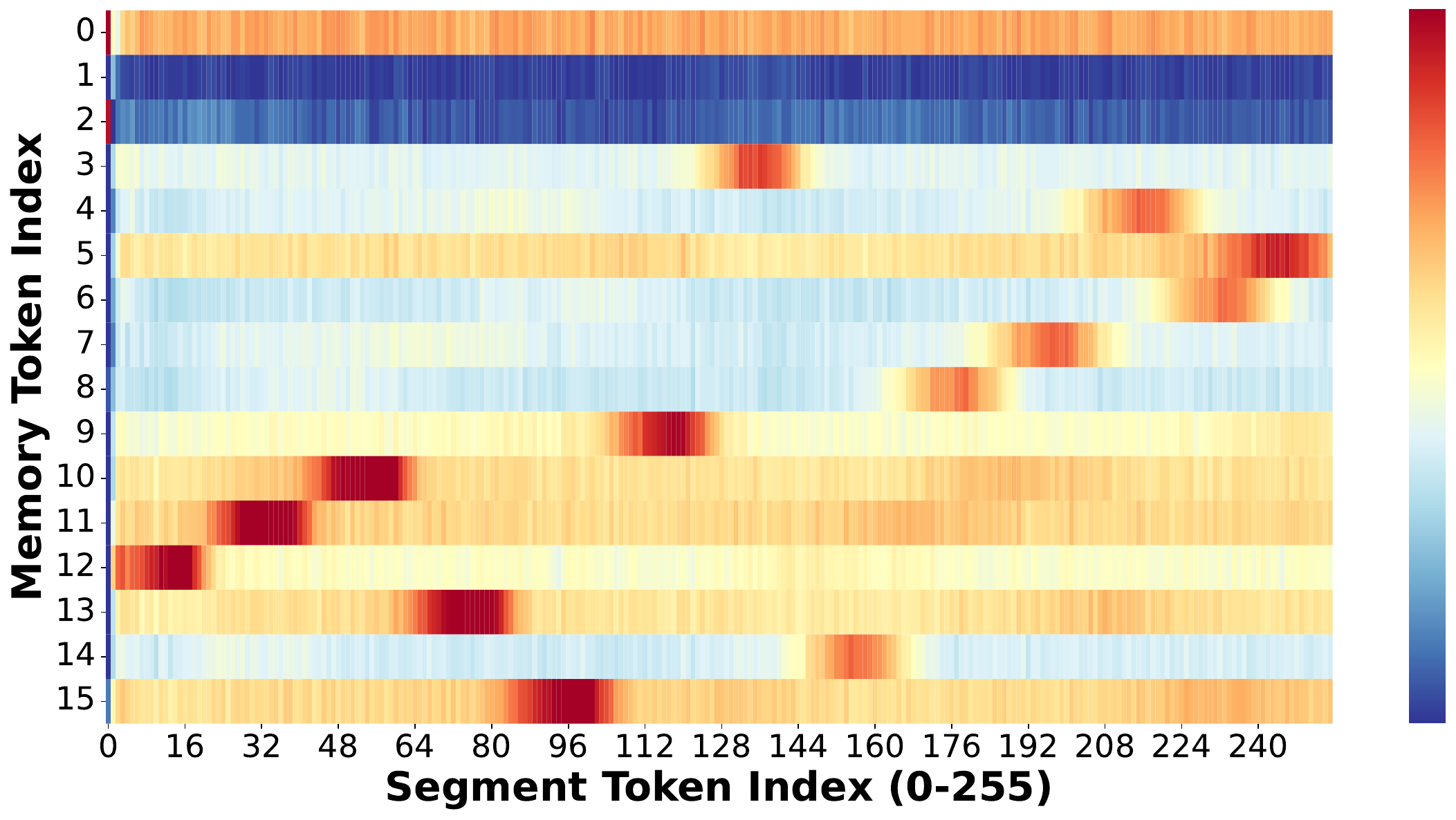}

        \caption{\textbf{pre-trained compressor}}
    \end{subfigure}
    
    \vspace{2pt} 
    
    \begin{subfigure}{0.4\linewidth}
        \centering
        \includegraphics[width=\textwidth]{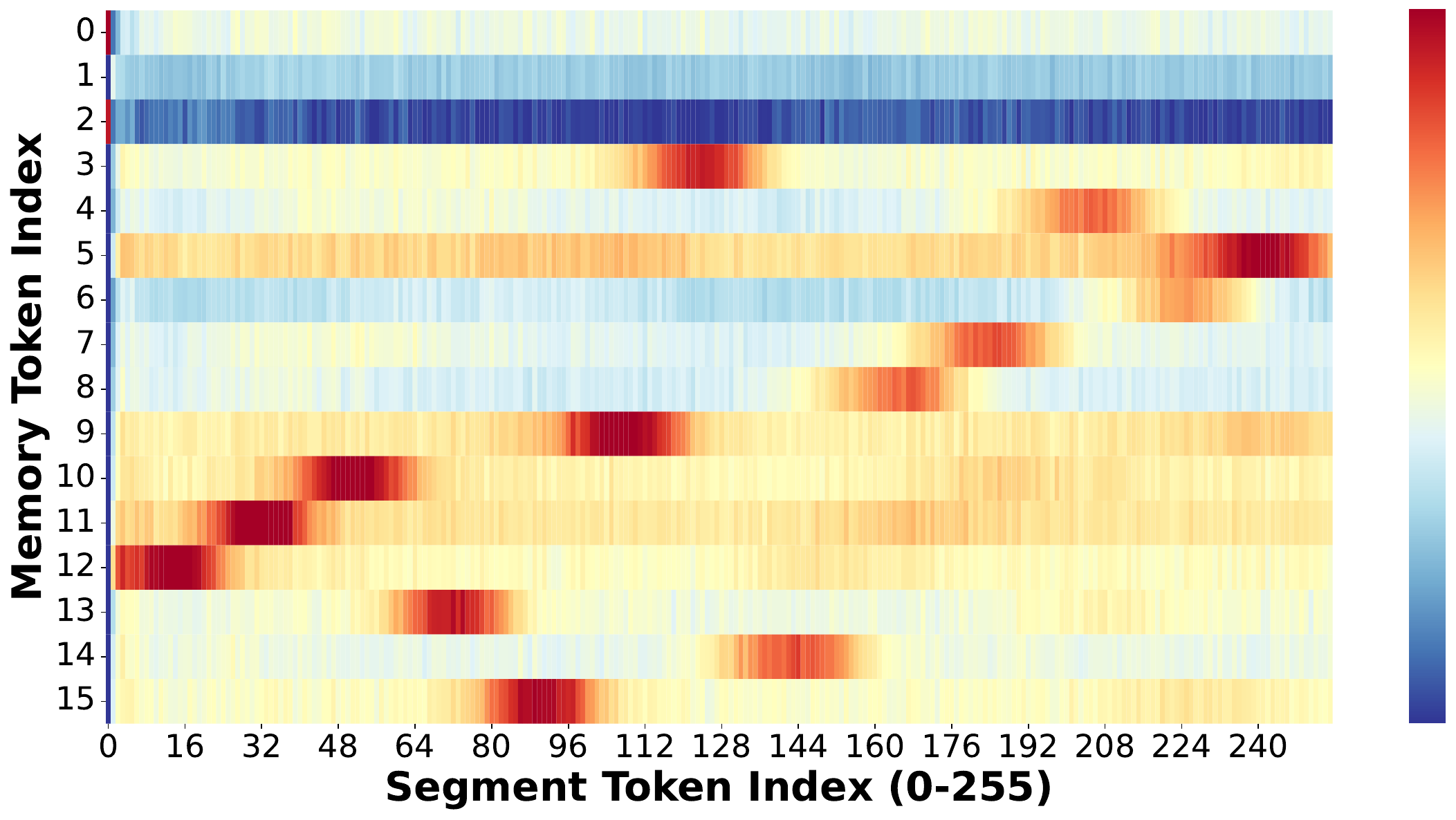}
        \caption{\textbf{fine-tuned compressor}}
    \end{subfigure}
    
    \caption{Heatmap visualization of the cosine similarity between memory embeddings (PCC) and original token embeddings on the Natural Questions dataset. Red indicates higher similarity, while blue represents lower similarity.}
    \label{fig.a.pcc.n}
\end{figure*}

We similarly conducted visualization experiments on the spatial specialization phenomenon of PCC across other datasets, as shown in Figure \ref{fig.a.pcc.attn.s}, Figure \ref{fig.a.pcc.s}, Figure \ref{fig.a.pcc.h}, Figure \ref{fig.a.pcc.attn.a}, Figure \ref{fig.a.pcc.a}, Figure \ref{fig.a.pcc.attn.n} and Figure \ref{fig.a.pcc.n}, we observe that this spatial specialization phenomenon is dataset-independent, with texts across different datasets exhibiting consistent spatial specialization patterns. Furthermore, the features become more pronounced as the number of samples in the dataset increases. However, in the case of the AdversarialQA dataset, the statistical regularity is less evident. This is primarily because the samples are generally short, with the majority having a length of less than 256 tokens.



\subsection{OURS (\sys)}\label{app:pic}
\begin{figure*}
    \centering

    \captionsetup[subfigure]{font=normalsize}

    \begin{subfigure}{0.4\linewidth}
        \centering

        \includegraphics[width=\textwidth]{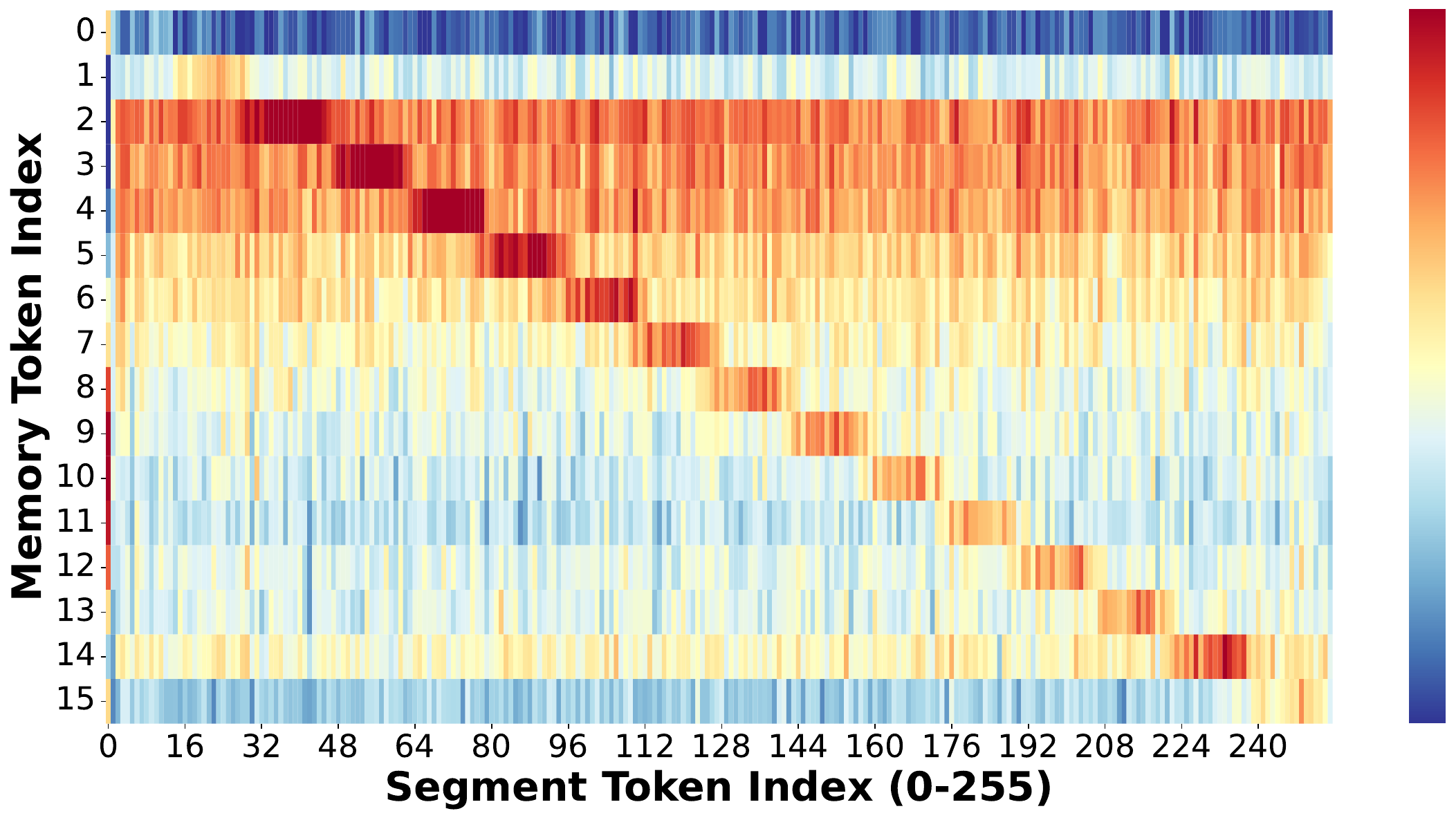}

        \caption{\textbf{pre-trained compressor}}
    \end{subfigure}
    
    \vspace{2pt} 
    
    \begin{subfigure}{0.4\linewidth}
        \centering
        \includegraphics[width=\textwidth]{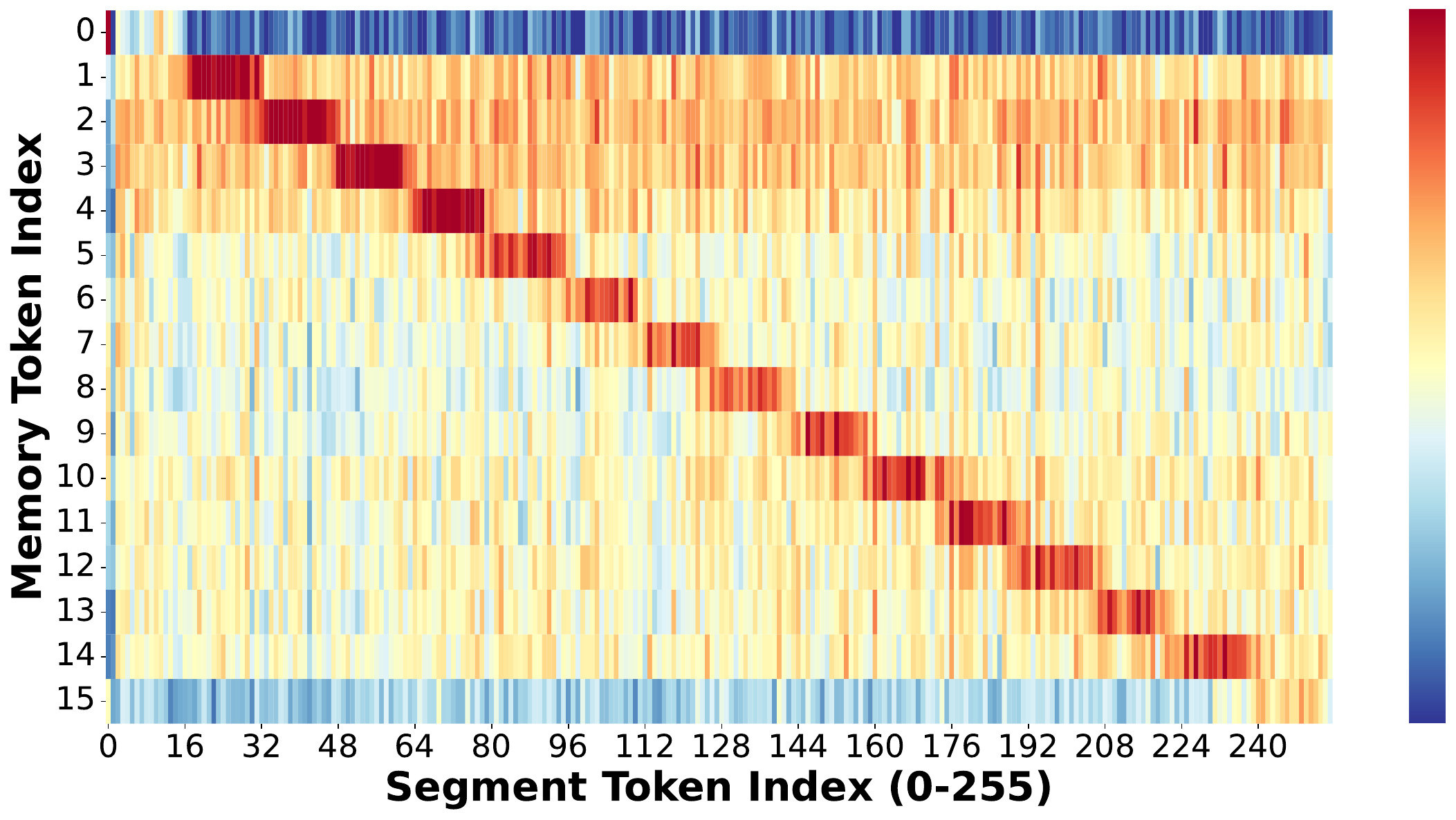}
        \caption{\textbf{fine-tuned compressor}}
    \end{subfigure}
    
    \caption{Heatmap visualization of the cosine similarity between memory embeddings (\sys) and original token embeddings on the SQuAD dataset. Red indicates higher similarity, while blue represents lower similarity.}
    \label{fig.a.pic.s}
\end{figure*}

\begin{figure*}
    \centering
    \captionsetup[subfigure]{font=normalsize}

    \begin{subfigure}{0.4\linewidth}
        \centering

        \includegraphics[width=\textwidth]{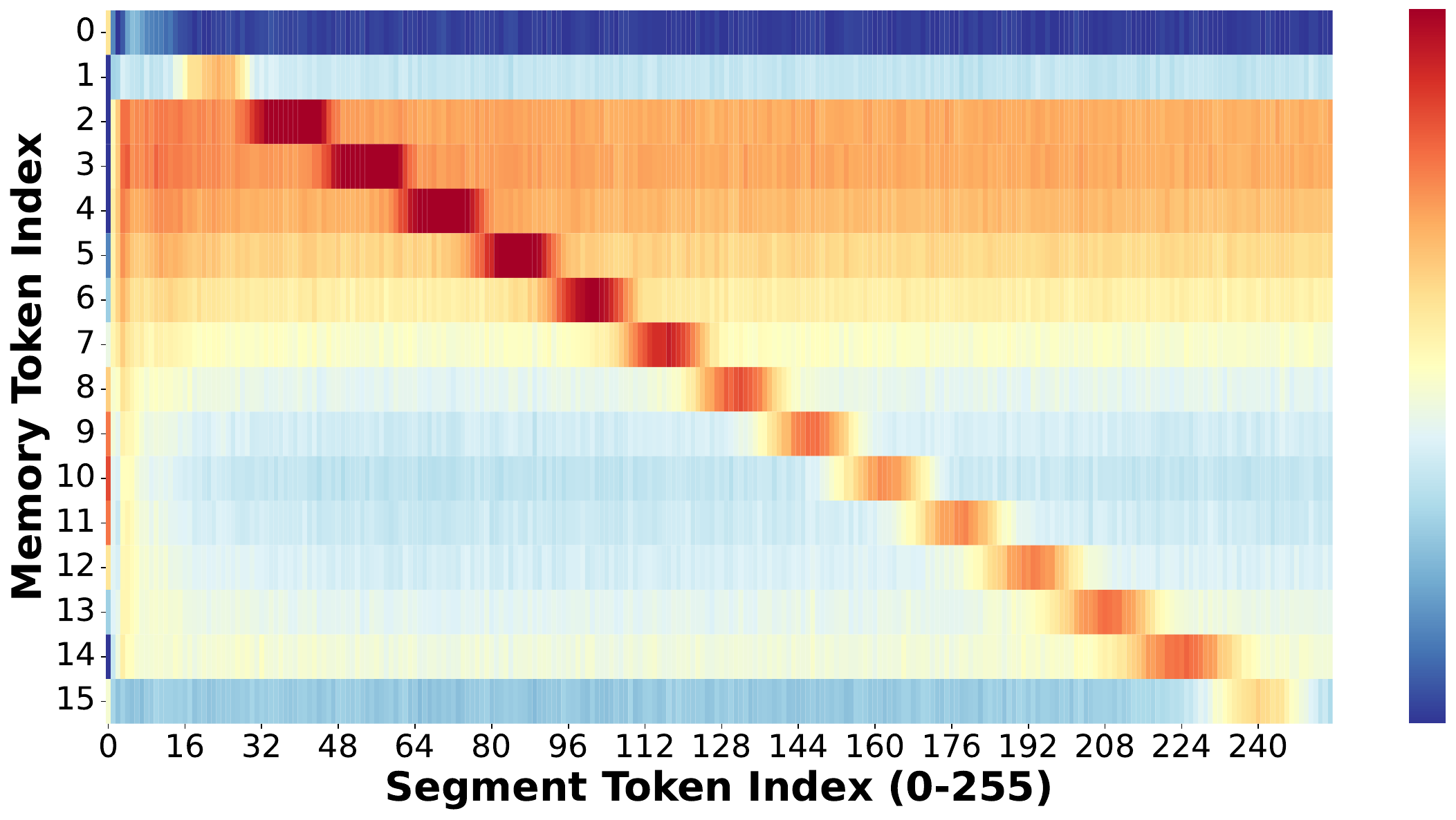}

        \caption{\textbf{pre-trained compressor}}
    \end{subfigure}
    
    \vspace{2pt} 
    
    \begin{subfigure}{0.4\linewidth}
        \centering
        \includegraphics[width=\textwidth]{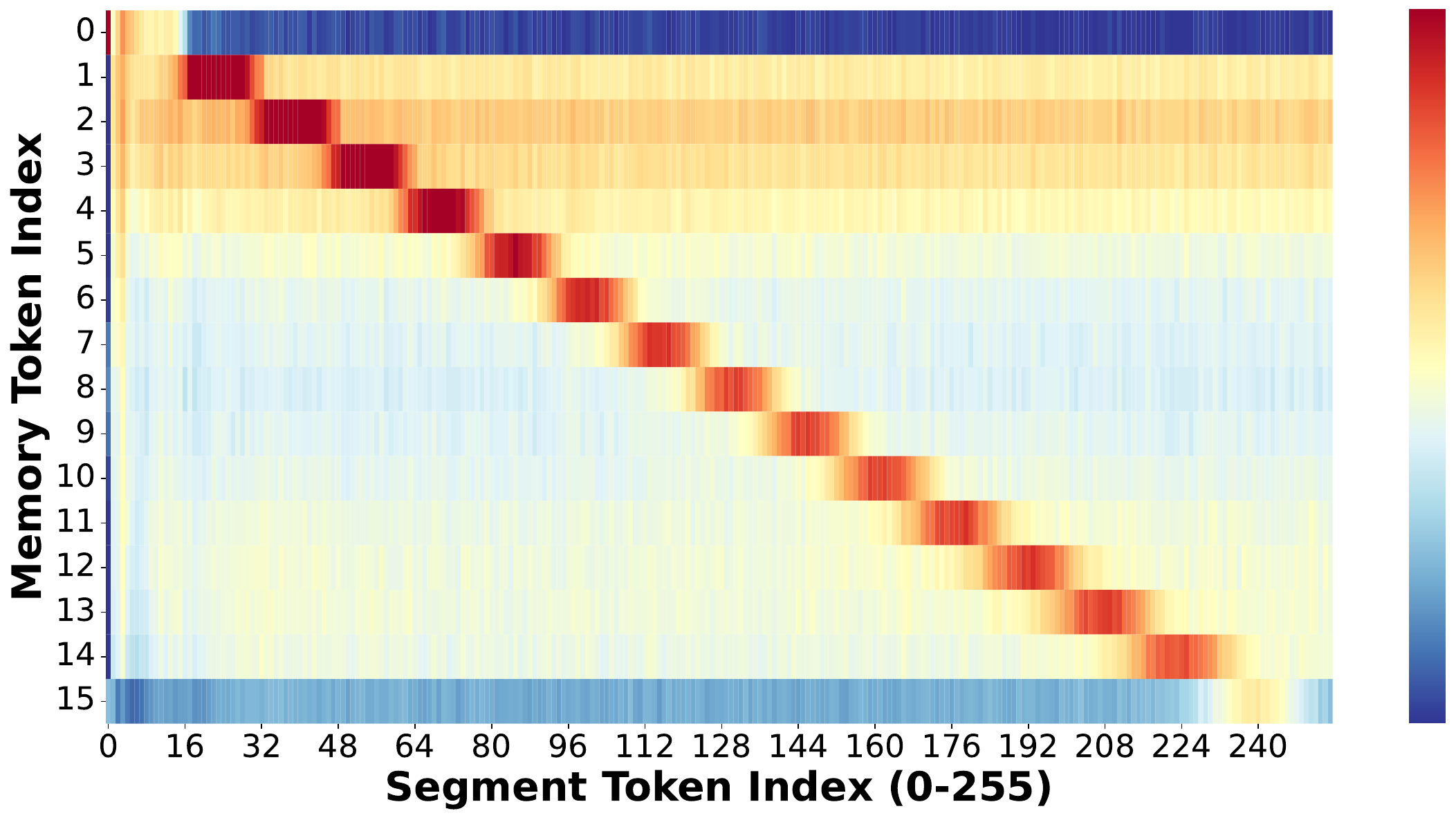}
        \caption{\textbf{fine-tuned compressor}}
    \end{subfigure}
    
    \caption{Heatmap visualization of the cosine similarity between memory embeddings (\sys) and original token embeddings on the HotpotQA dataset. Red indicates higher similarity, while blue represents lower similarity.}
    \label{fig.a.pic.h}
\end{figure*}

\begin{figure*}
    \centering

    \captionsetup[subfigure]{font=normalsize}

    \begin{subfigure}{0.4\linewidth}
        \centering

        \includegraphics[width=\textwidth]{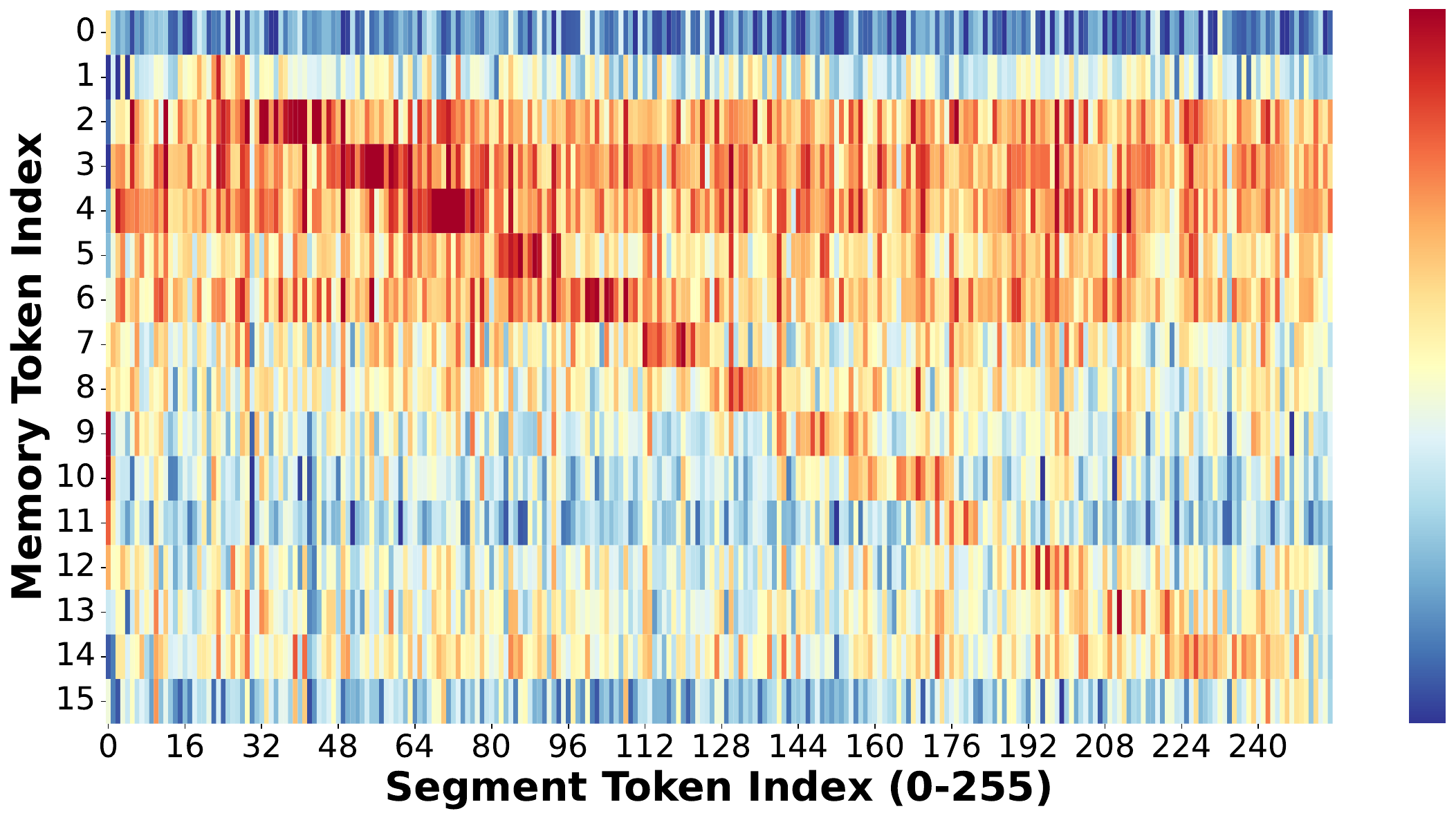}

        \caption{\textbf{pre-trained compressor}}
    \end{subfigure}
    
    \vspace{2pt} 
    
    \begin{subfigure}{0.4\linewidth}
        \centering
        \includegraphics[width=\textwidth]{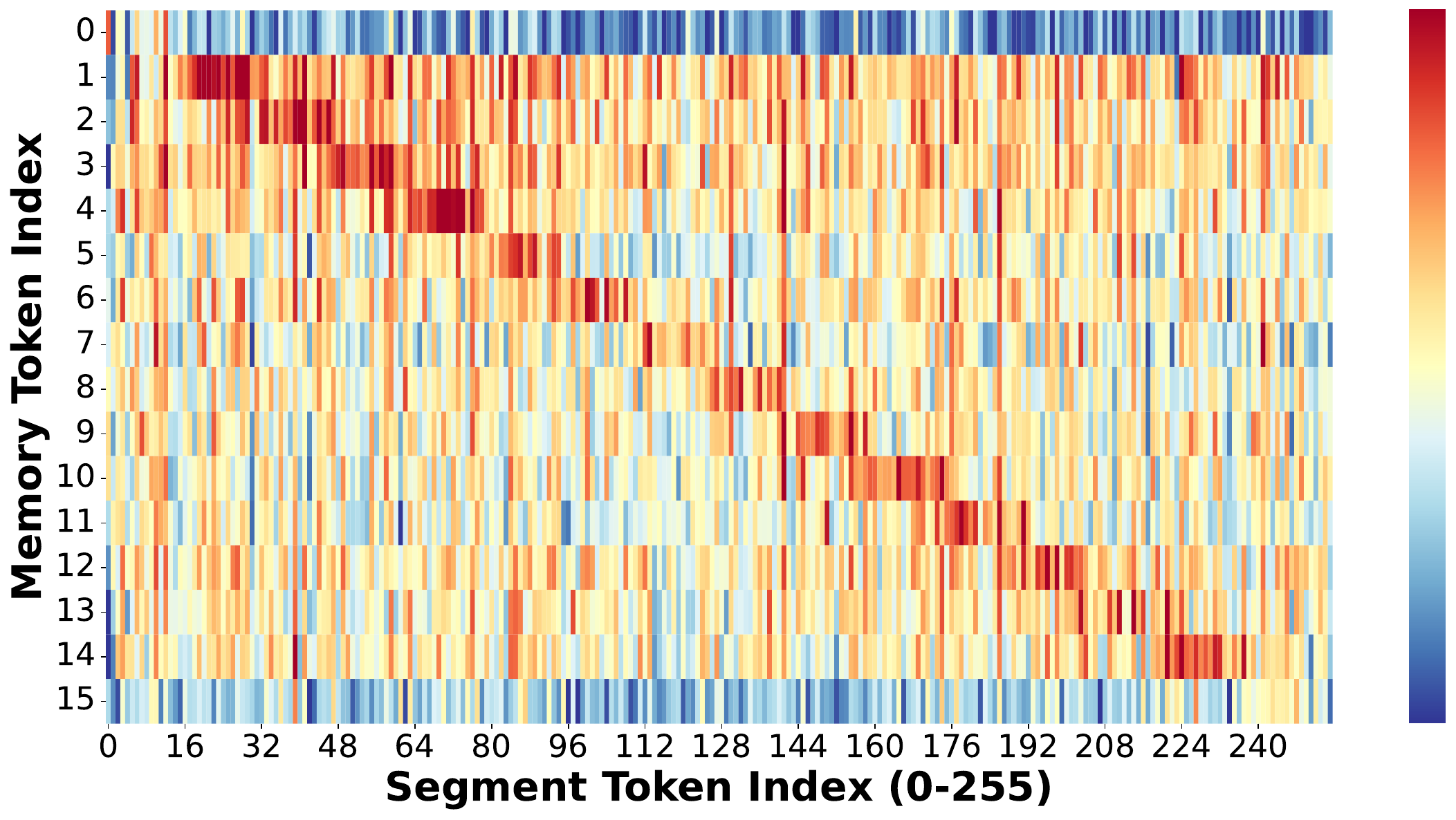}
        \caption{\textbf{fine-tuned compressor}}
    \end{subfigure}
    
    \caption{Heatmap visualization of the cosine similarity between memory embeddings (\sys) and original token embeddings on the AdversarialQA dataset. Red indicates higher similarity, while blue represents lower similarity.}
    \label{fig.a.pic.a}
\end{figure*}

\begin{figure*}
    \centering

    \captionsetup[subfigure]{font=normalsize}

    \begin{subfigure}{0.4\linewidth}
        \centering

        \includegraphics[width=\textwidth]{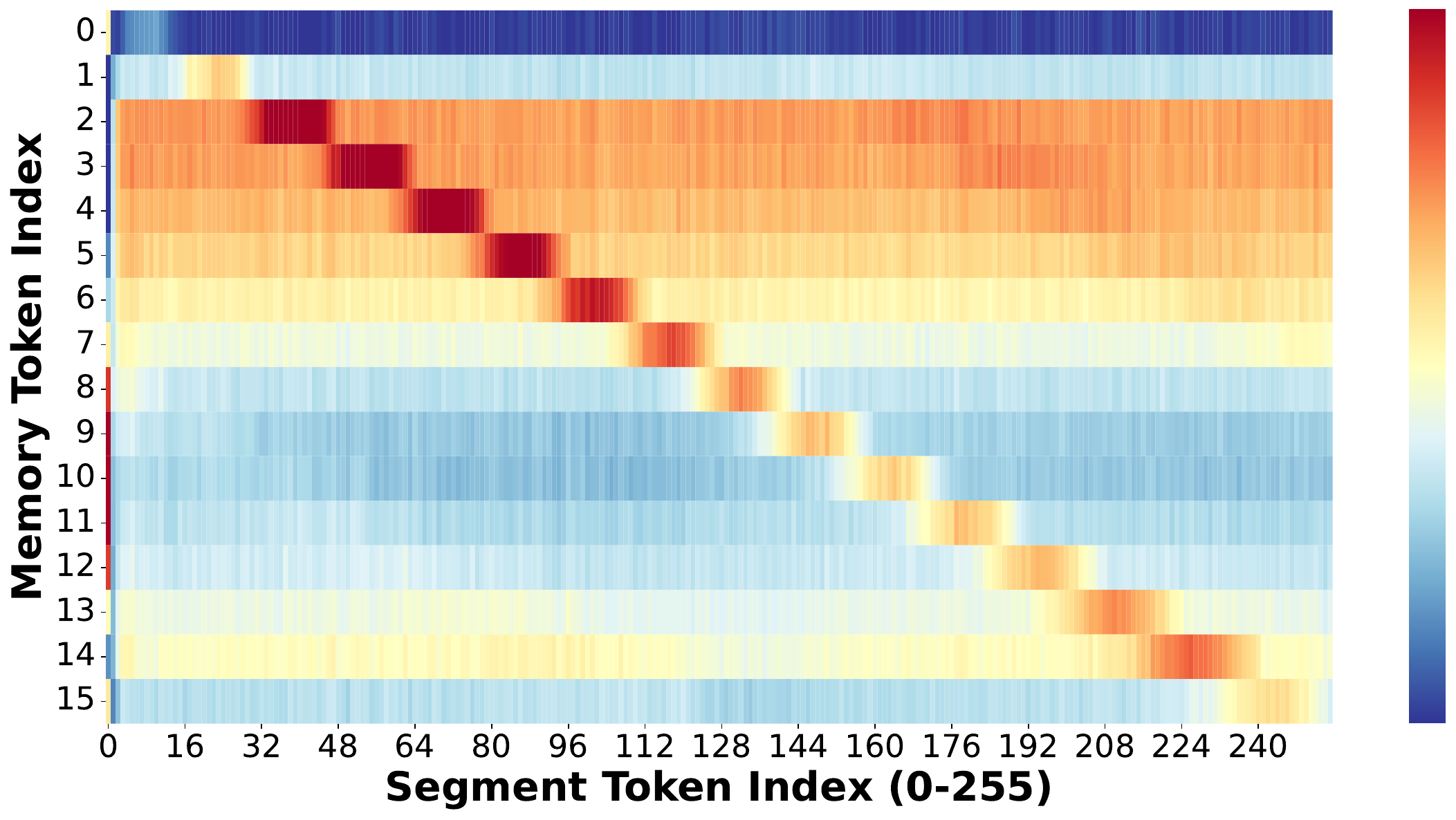}

        \caption{\textbf{pre-trained compressor}}
    \end{subfigure}

    \vspace{2pt} 
    
    \begin{subfigure}{0.4\linewidth}
        \centering
        \includegraphics[width=\textwidth]{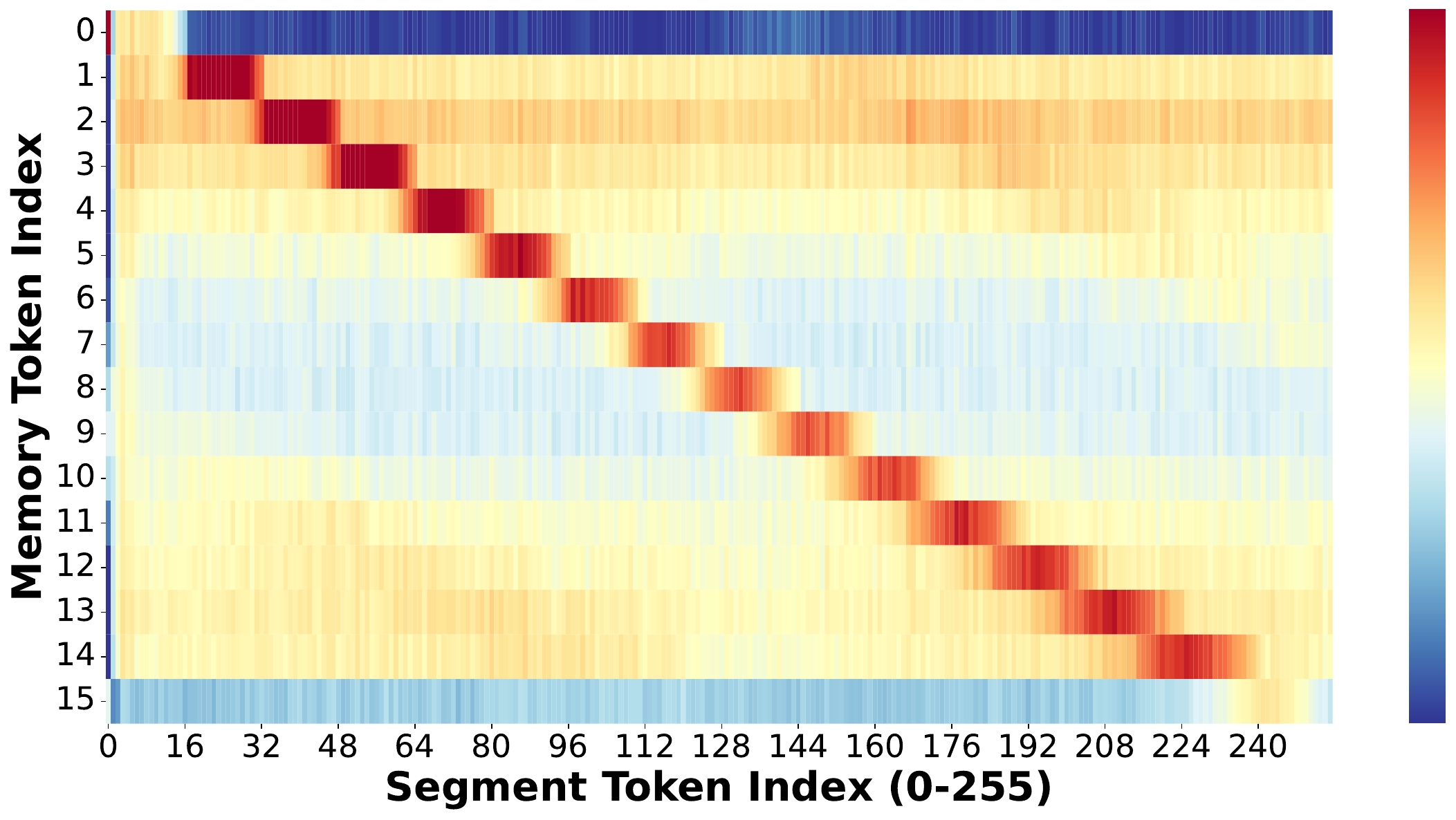}
        \caption{\textbf{fine-tuned compressor}}
    \end{subfigure}

    \caption{Heatmap visualization of the cosine similarity between memory embeddings (\sys) and original token embeddings on the Natural Questions dataset. Red indicates higher similarity, while blue represents lower similarity.}
    \label{fig.a.pic.n}
\end{figure*}

We similarly conducted visualization experiments on the spatial specialization phenomenon of our method across other datasets, as shown in Figure \ref{fig.a.pic.s}, Figure \ref{fig.a.pic.h}, Figure \ref{fig.a.pic.a}, and Figure \ref{fig.a.pic.n}, \sys~ displays a sharp and consistent diagonal pattern across all evaluated domains. This confirms that our block-wise causal mask effectively enforces the intended receptive field constraints, ensuring that each memory token is exclusively dedicated to compressing its corresponding local context chunk. This structural alignment is robust across different data distributions, validating the generalizability of our approach..

\section{Analysis of Cosine Similarity Between Embedding Pairs on Other Datasets}\label{app:m2m}
\begin{figure*}
    \centering

    \captionsetup[subfigure]{font=small}

    \begin{subfigure}{0.4\linewidth}
        \centering

        \includegraphics[width=\textwidth]{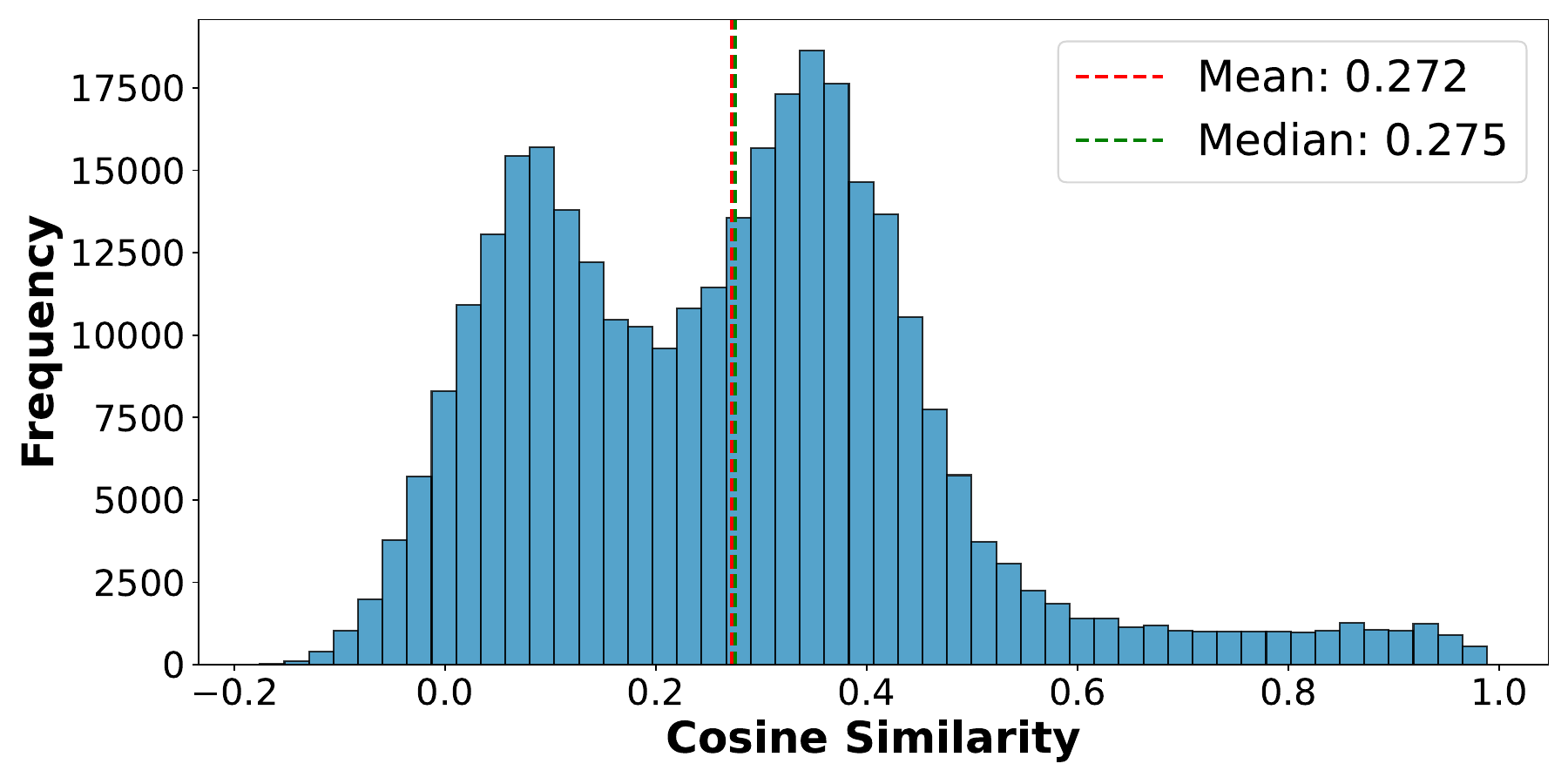}

        \caption{\textbf{PCC}}
    \end{subfigure}

    \vspace{2pt}

    \begin{subfigure}{0.4\linewidth}
        \centering
        \includegraphics[width=\textwidth]{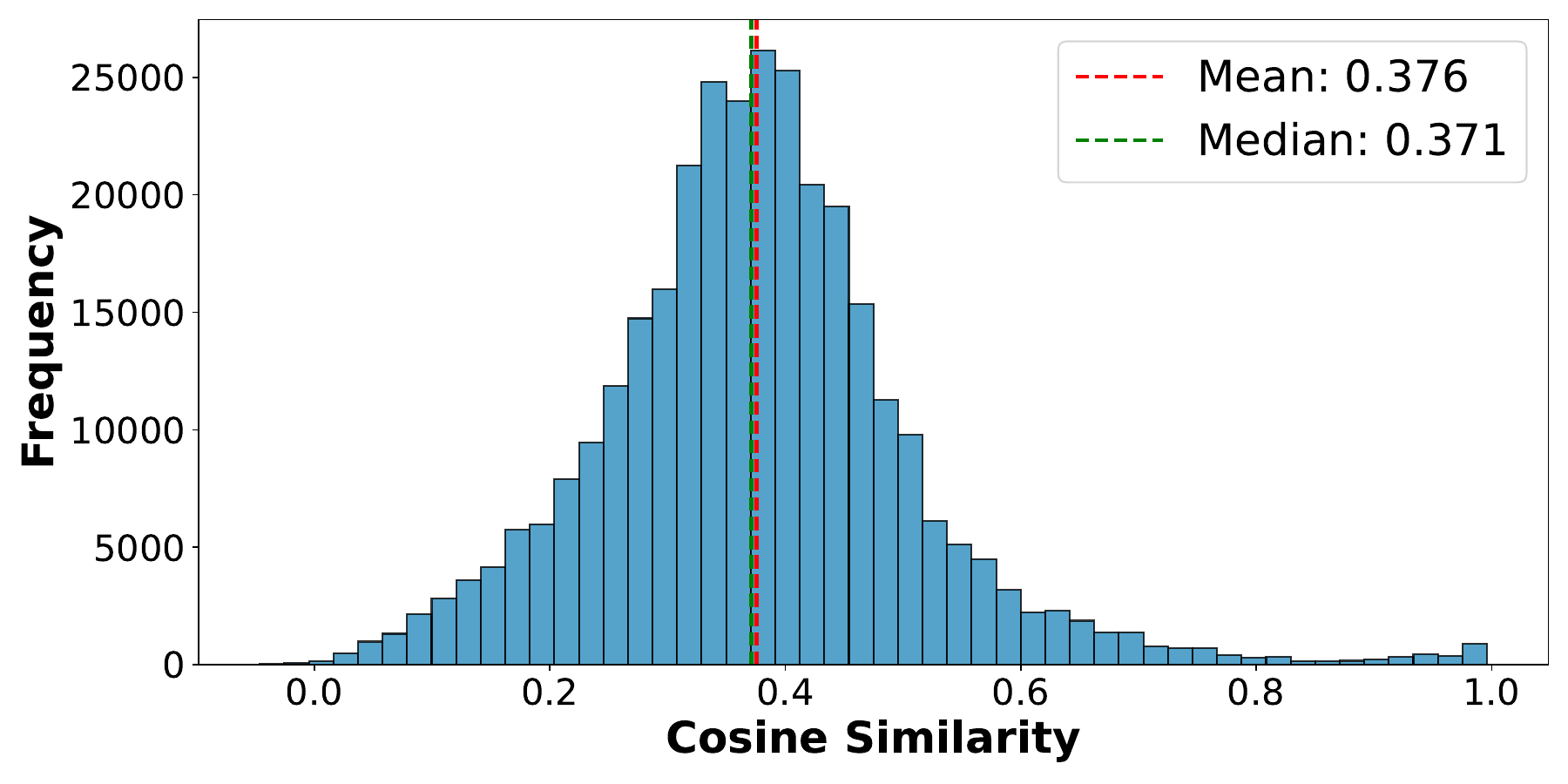}
        \caption{\textbf{OURS (\sys)}}
    \end{subfigure}

    \caption{Visualization of pairwise cosine similarity distributions between memory embeddings within the same memory slot on the SQuAD dataset. The compressors used are fine-tuned on downstream QA tasks.}
    \label{fig.a.m2m.s}
\end{figure*}

\begin{figure*}
    \centering
    \captionsetup[subfigure]{font=small}

    \begin{subfigure}{0.4\linewidth}
        \centering

        \includegraphics[width=\textwidth]{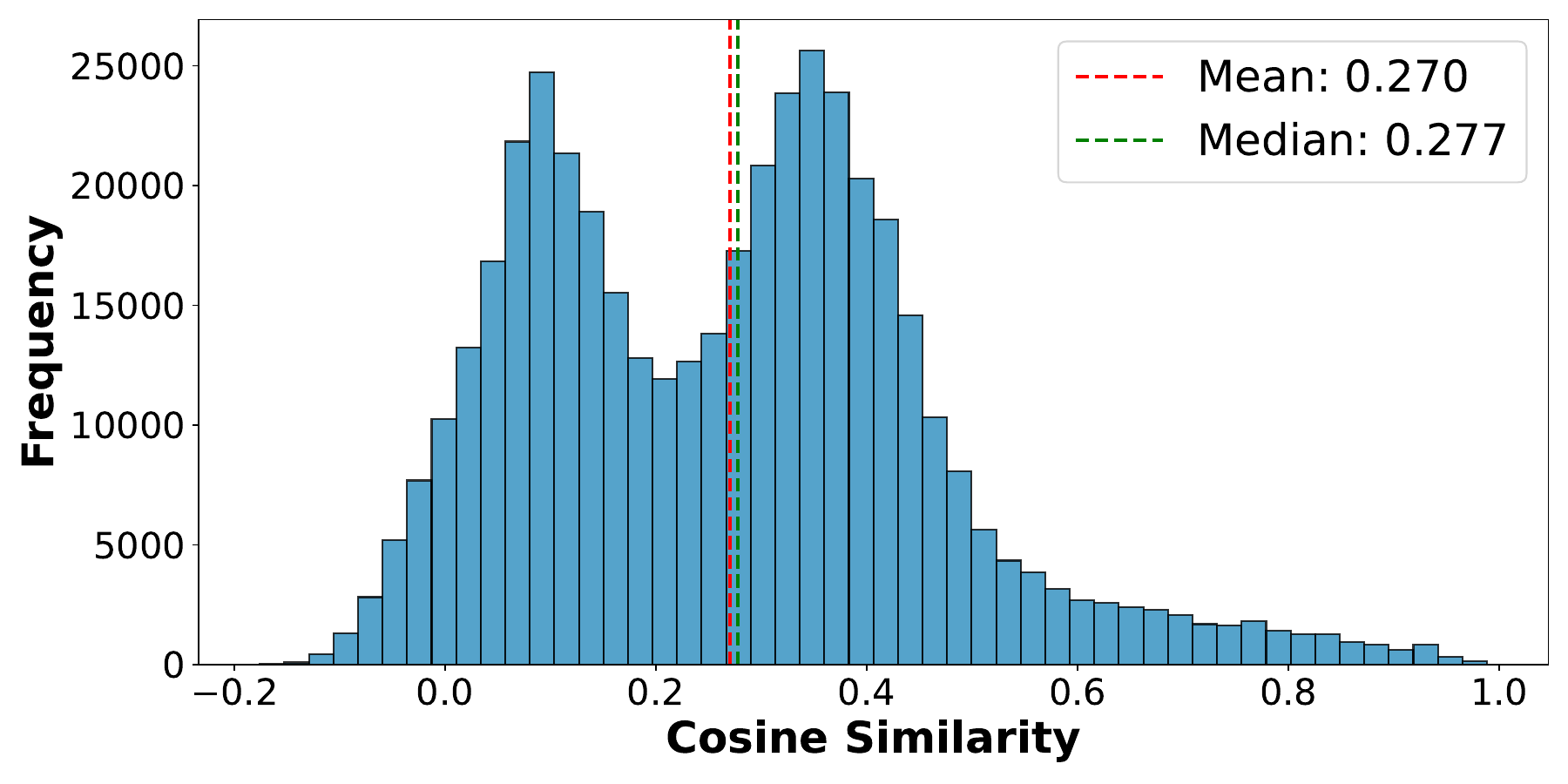}

        \caption{\textbf{PCC}}
    \end{subfigure}

    \vspace{2pt} 
    
    \begin{subfigure}{0.4\linewidth}
        \centering
        \includegraphics[width=\textwidth]{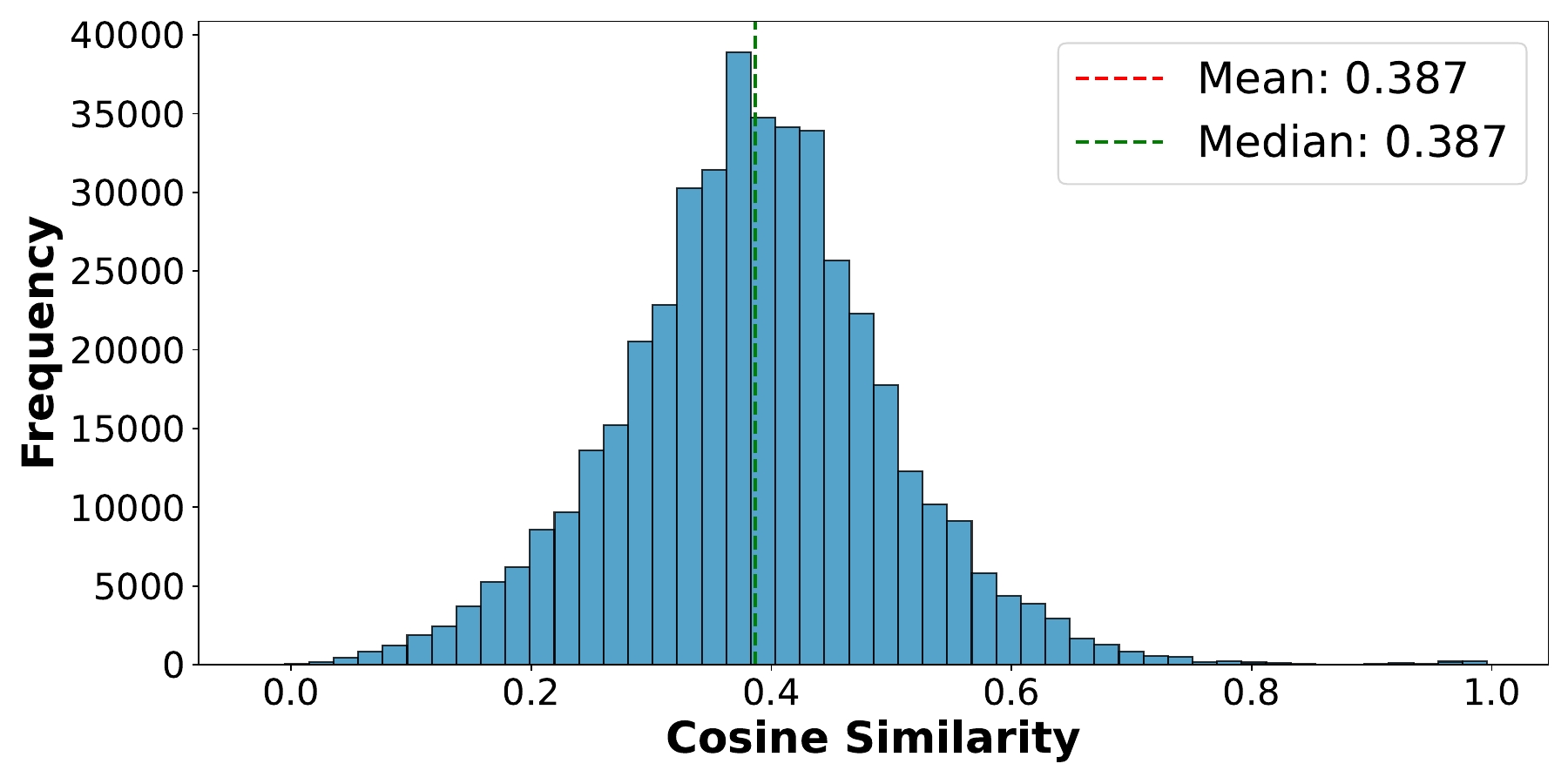}
        \caption{\textbf{OURS (\sys)}}
    \end{subfigure}

    \caption{Visualization of pairwise cosine similarity distributions between memory embeddings within the same memory slot on the AdversarialQA dataset. The compressors used are fine-tuned on downstream QA tasks.}
    \label{fig.a.m2m.a}
\end{figure*}

\begin{figure*}
    \centering

    \captionsetup[subfigure]{font=small}

    \begin{subfigure}{0.4\linewidth}
        \centering

        \includegraphics[width=\textwidth]{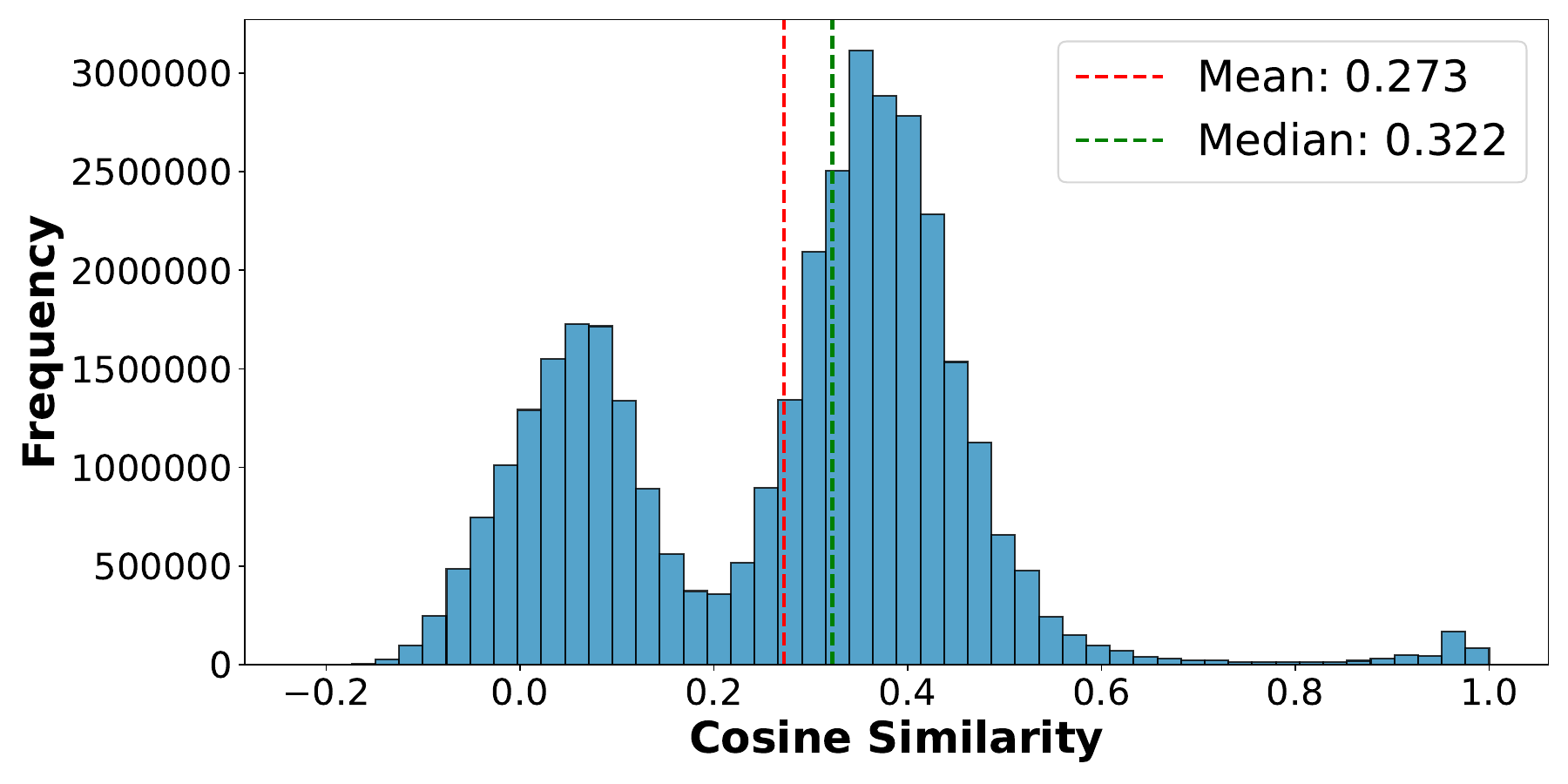}
 
        \caption{\textbf{PCC}}
    \end{subfigure}

    \vspace{2pt}

    \begin{subfigure}{0.4\linewidth}
        \centering
        \includegraphics[width=\textwidth]{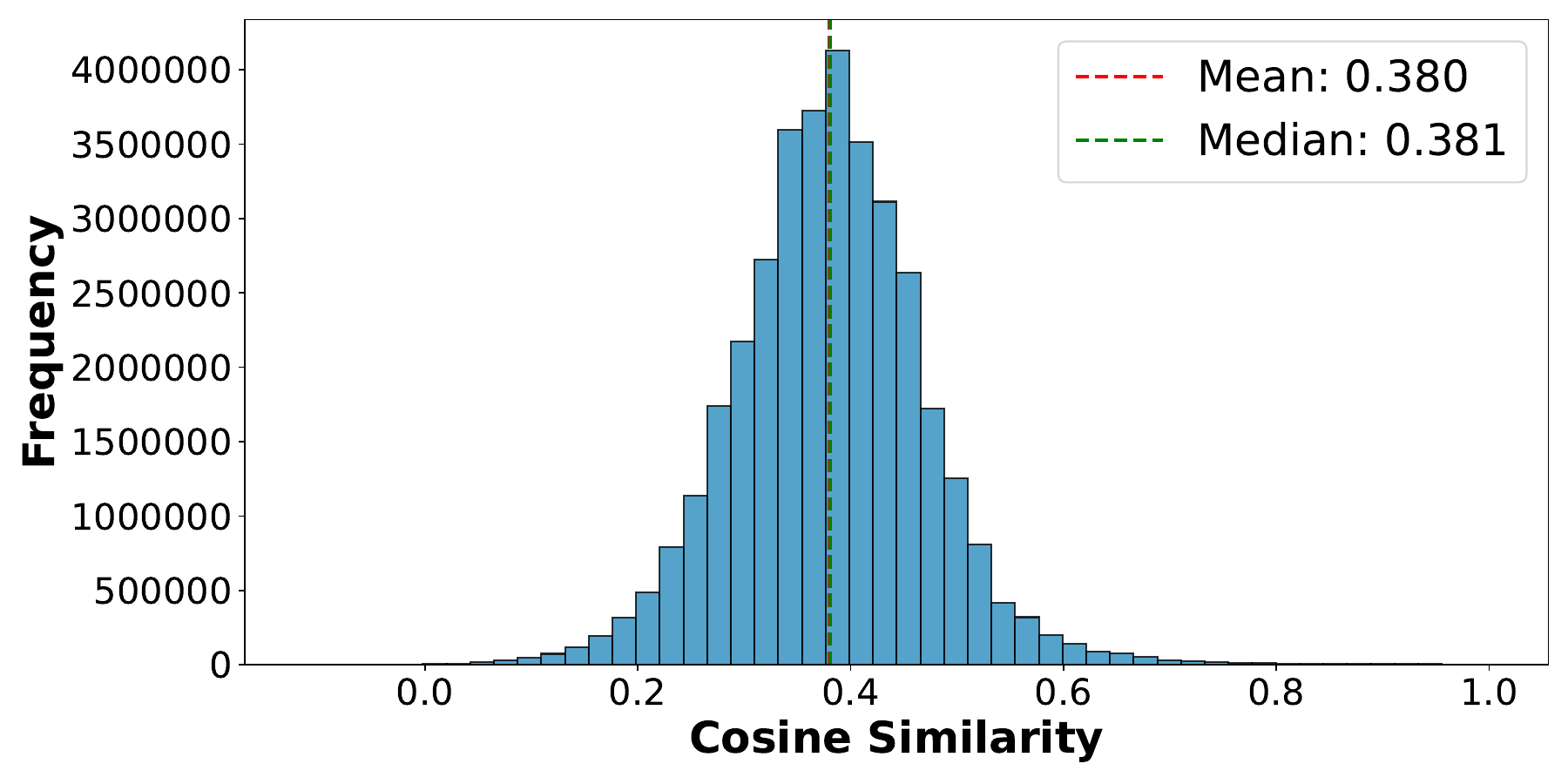}
        \caption{\textbf{OURS (\sys)}}
    \end{subfigure}

    \caption{Visualization of pairwise cosine similarity distributions between memory embeddings within the same memory slot on the Natural Questions dataset. The compressors used are fine-tuned on downstream QA tasks.}
    \label{fig.a.m2m.n}
\end{figure*}

We also analyzed the cosine similarity between memory embeddings on other datasets; see Figure \ref{fig.a.m2m.s}, Figure \ref{fig.a.m2m.a}, and Figure \ref{fig.a.m2m.n}, the pairwise cosine similarity distributions for PCC consistently exhibit a bimodal pattern with a heavy tail of high similarity, indicating significant redundancy and potential mode collapse among memory tokens. Conversely, \sys~ maintains a distribution centered around zero with lower variance across all datasets. This suggests that our method produces more orthogonal and information-dense memory representations, effectively utilizing the memory budget by minimizing inter-token redundancy..

\section{Data Efficiency Analysis on Other Compression Rate}
\begin{figure*}
    \centering

    \captionsetup[subfigure]{font=normalsize}

    \begin{subfigure}{0.5\linewidth}
        \centering

        \includegraphics[width=\textwidth]{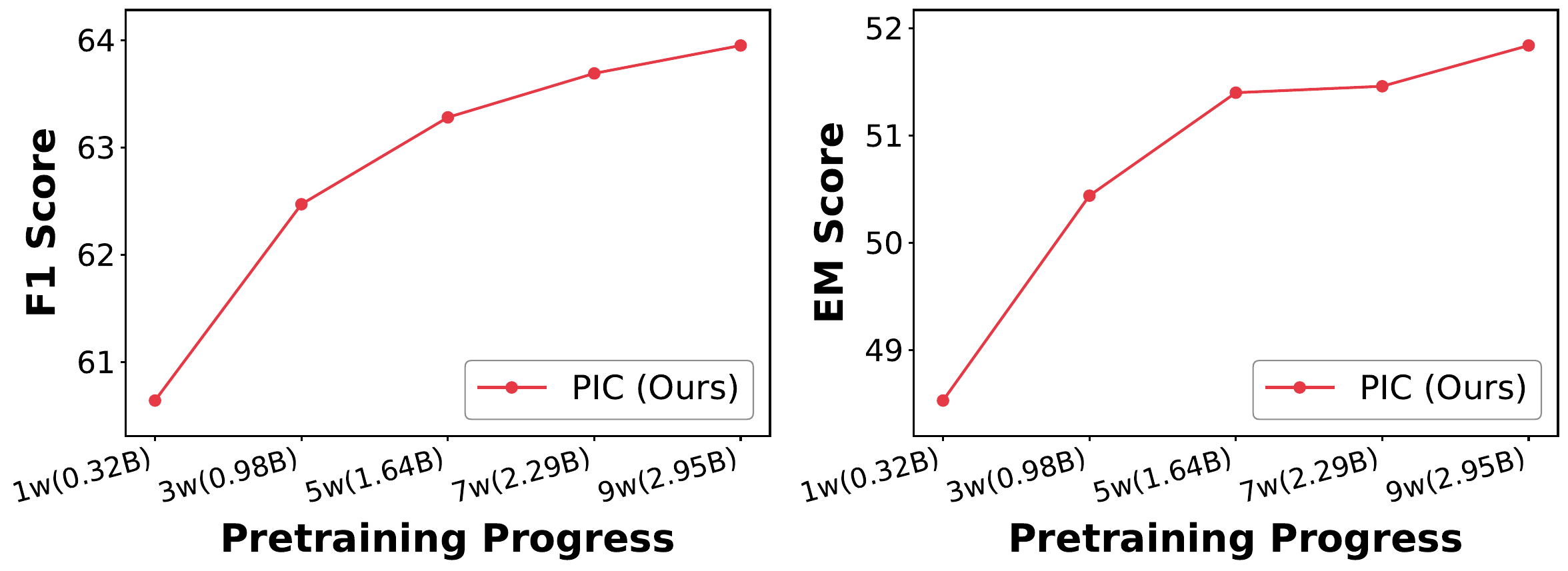}

        \caption{\textbf{4x Compressor}}
    \end{subfigure}
    
    \vspace{2pt} 
    
    \begin{subfigure}{0.5\linewidth}
        \centering
        \includegraphics[width=\textwidth]{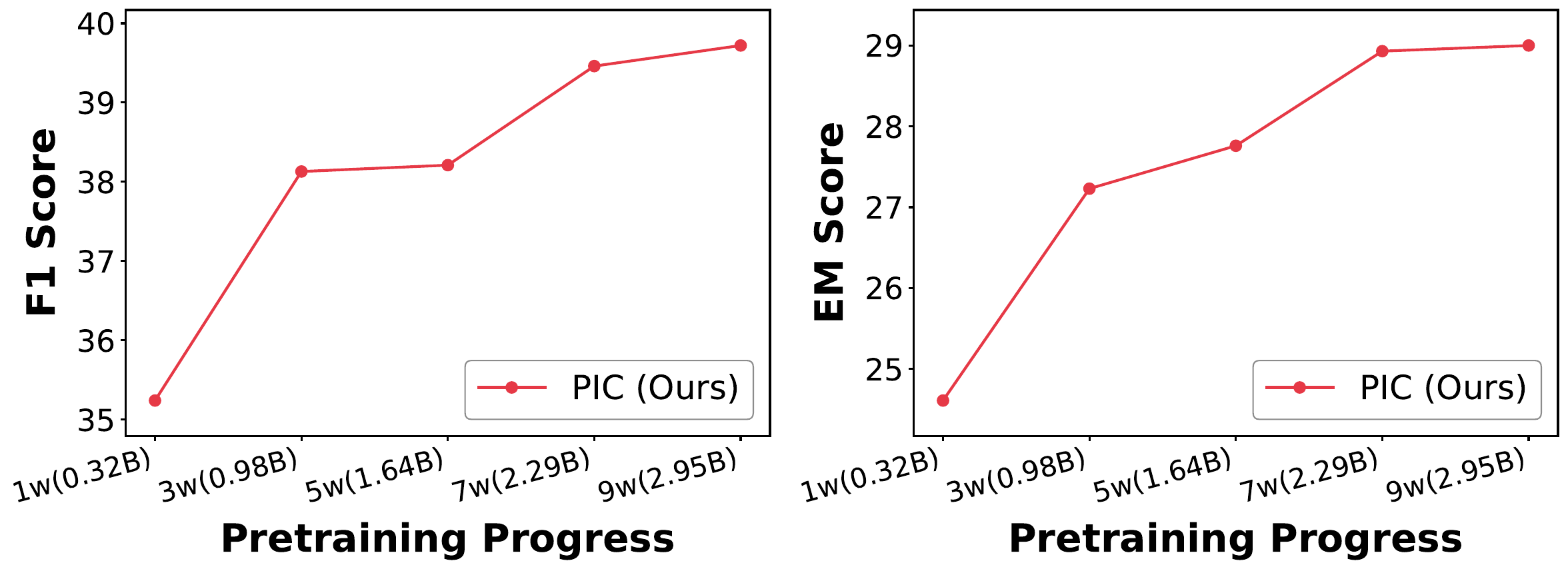}
        \caption{\textbf{64x Compressor}}
    \end{subfigure}
    
    \caption{Relationship between Compressor Performance on Downstream QA Tasks and Pre-training Steps. The x-axis represents the number of tokens corresponding to the pre-training steps (converted to 1 epoch equivalent), and the y-axis represents the EM or F1 score.}
    \label{fig.a.4x.64x}
\end{figure*}

We further investigated the data efficiency of our method under different compression regimes. Figure \ref{fig.a.4x.64x} presents the performance trajectories for 4x and 64x compression rates. Consistent with the 16x results discussed in the main text, \sys~ demonstrates a superior learning curve. For both compression ratios, our method achieves higher downstream performance (EM and F1 scores) at earlier pre-training steps compared to the baseline, confirming that the block-wise constraint accelerates the acquisition of effective compression capabilities regardless of the compression granularity.

\section{Complete Results of Ablation Study}
Table \ref{tab.a.1} provides the comprehensive breakdown of our ablation study. Across all four datasets (SQuAD, HotPotQA, AdversarialQA, and NQ), the \sys~ model equipped with the block-wise causal mask consistently outperforms the variant using a full causal mask (PCC setting). The performance gap is evident in both F1 and EM metrics, statistically validating that the structured inductive bias introduced by our masking strategy is the primary driver of the observed improvements, rather than the model architecture alone.
\begin{table*}
    \centering
    \caption{Ablation Study: The table presents the performance ( F1 and Exact Match (EM)) across four datasets: SQuAD, HotPotQA, AdversarialQA, and NQ, along with the average metrics. Our method is compared with the PCC baseline (using full causal mask). Best-performing results for each dataset are highlighted in bold.}
    \label{tab.a.1}    
    \begin{tabular*}{\linewidth}{@{\extracolsep{\fill}}cccccccccccc}
        \toprule
         & \multicolumn{2}{c}{SQuAD} & \multicolumn{2}{c}{HotPotQA} & \multicolumn{2}{c}{AdversarialQA} & \multicolumn{2}{c}{NQ} & \multicolumn{2}{c}{Average}\\
        \cmidrule(lr){2-3} \cmidrule(lr){4-5} \cmidrule(lr){6-7} \cmidrule(lr){8-9} \cmidrule(lr){10-11}
        Methods & $F_1$ & EM & $F_1$ & EM & $F_1$ & EM & $F_1$ & EM & $F_1$ & EM &\\
        \midrule
        OURS (\sys)     & \textbf{59.41}   & \textbf{39.86}   & \textbf{41.05}   & \textbf{33.24}   & \textbf{39.98}   & \textbf{27.73}   & \textbf{67.61}   & \textbf{58.37}   & \textbf{52.01}  & \textbf{39.80} \\
        w/ full casual mask (PCC) & 55.37   & 36.12   & 36.07   & 28.24   & 37.16   & 24.93   & 66.60   & 57.93   & 48.80  & 36.81 \\
        \bottomrule
    \end{tabular*}
\end{table*}

\end{document}